\documentclass{article}
\usepackage{iclr2027_conference,times}
\iclrfinalcopy 
\usepackage[most]{tcolorbox}
\usepackage{caption}
\usepackage{amsmath}
\usepackage{amsfonts}
\usepackage{graphicx}
\usepackage{float}
\usepackage{algorithmic}
\usepackage{algorithm}
\usepackage{array}
\usepackage{textcomp}
\usepackage{stfloats}
\usepackage{verbatim}
\usepackage{etoolbox}

\usepackage{makecell}
\usepackage{multicol}
\usepackage{wrapfig}
\usepackage{needspace}
\usepackage{times}
\usepackage{siunitx}
\usepackage{gensymb}
\usepackage{booktabs}
\usepackage{multirow}
\usepackage{xspace}

\usepackage{xcolor}
\usepackage{todonotes}
\usepackage{adjustbox}
\usepackage{placeins}
\usepackage{pdflscape}

\usepackage{bbm}
\usepackage{tikz}
\usepackage{enumitem}

\usepackage{hyperref}
\hypersetup{
    colorlinks=true,
    linkcolor=black,
    urlcolor=black,
    citecolor=black
}

\title{Temporal Self-Imitation Learning}

\author{Yinsen Jia, Boyuan Chen\\
Duke University\\
\url{https://generalroboticslab.com/TSIL}
}

\begin{document}
\maketitle
\lhead{Preprint}

\begin{abstract}
Long-horizon policies trained with reinforcement learning can still achieve high return through inefficient interactions, while rare efficient behaviors discovered during training may be forgotten. We argue that temporal efficiency itself provides a source of self-supervision for reinforcement learning. We introduce \textbf{Temporal Self-Imitation Learning (TSIL)}, a reinforcement learning framework that mines temporally efficient successful trajectories generated during learning and converts them into reusable supervision for future policy improvement. TSIL progressively refines learning using configuration-conditioned adaptive temporal targets derived from fast successful trajectories, while preserving and replaying efficient behaviors through efficiency-weighted self-imitation learning. Across 30 long-horizon tasks spanning robot manipulation and interactive navigation, TSIL consistently improves task success rates, learning efficiency, behavioral efficiency, and robustness to unstable training conditions.
\end{abstract}

\begin{figure}[H]
    \centering
    \includegraphics[width=\textwidth,trim=0 32bp 0 27bp,clip]{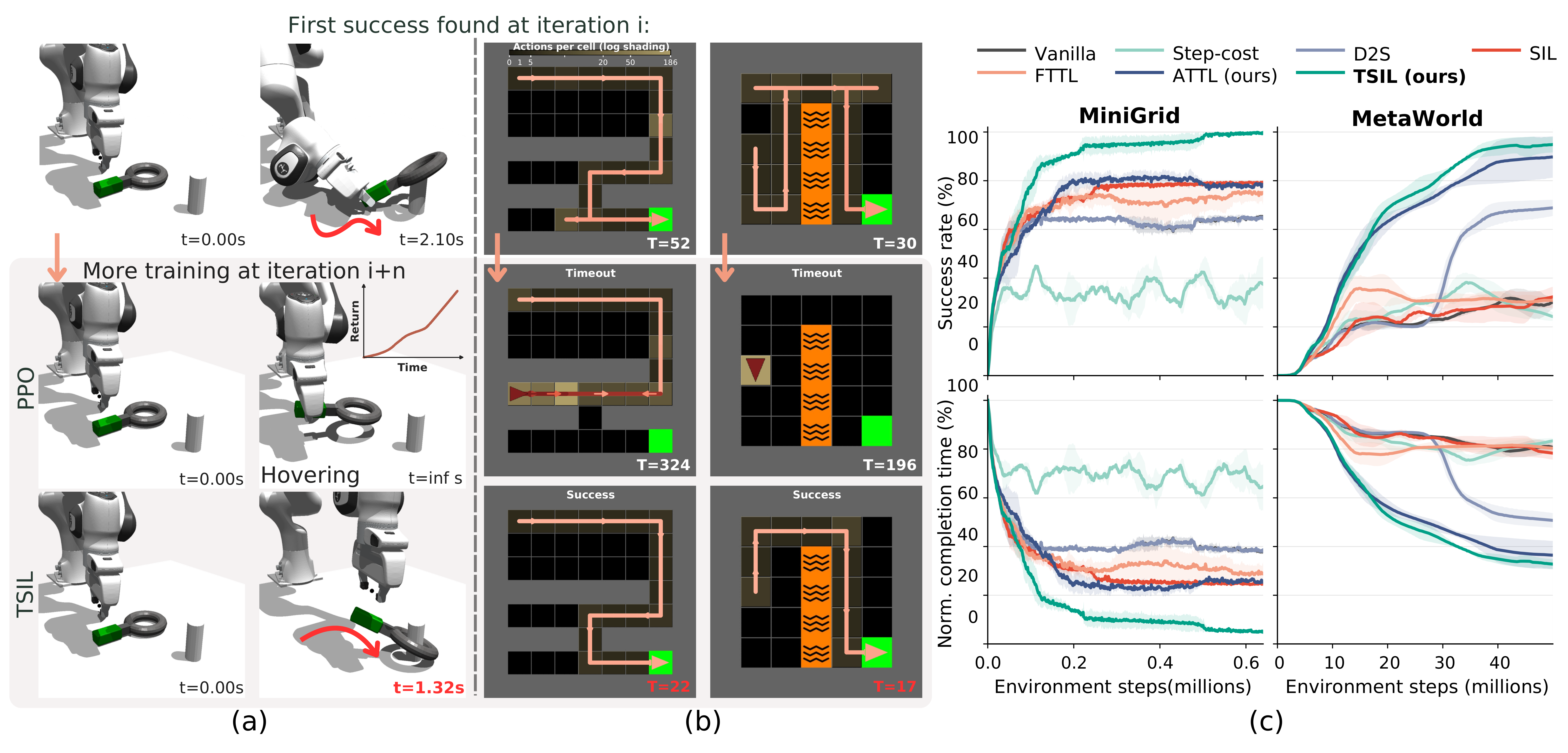}
    \caption{\textbf{TSIL focuses on three challenges in long-horizon RL.} \textbf{(a) Distraction:} In a dense-reward environment, the policy may exploit shaping rewards and drift away from previously discovered successful behavior. TSIL instead regularizes learning toward a temporal target derived from the previous fastest success. \textbf{(b) Forgetting:} In a sparse-reward environment, the policy may fail to retain successful behavior after discovering it. TSIL helps the policy revisit actions from previous fast successes. \textbf{(c) Efficiency:} TSIL continually refines fast successful behavior, improving both training and behavioral efficiency for task completion.}
    \label{fig:teaser}
    \vspace{-1em}
\end{figure}


\section{Introduction}
\label{sec:introduction}

Motor learning in humans and animals progressively refines memorized behavior toward more efficient and reliable movement. Through repeated interaction and reinforcement, biological systems learn to suppress unnecessary motion, reduce behavioral variability, and discover increasingly direct strategies for achieving a goal~\citep{Krakauer2019Motor,Vassiliadis2021Reward}. Skilled behavior therefore emerges not only from eventual task success, but from refining success into efficient and reproducible interaction patterns.

In contrast, long-horizon reinforcement learning can struggle to turn a discovered success into reliable future behavior. Reward objectives may imperfectly reflect efficient task completion, so useful behavior may not be reinforced by subsequent updates even when it remains represented in the data. Continued policy updates also shift the distribution of collected trajectories, leading rare successes to disappear from later training. Together, these issues can prevent a discovered success from becoming an efficient and reliable behavior.

Specifically, one problem with dense rewards is \textbf{distraction}. Long-horizon manipulation policies are commonly trained with shaping rewards alongside task-success rewards~\citep{gu2017deep,Schulman2017PPO,yu2020meta}. A robot may exploit high-reward intermediate states, remain stuck near partially solved configurations, or take much longer manipulation paths before eventually completing the task. Such behavior can achieve high return without solving the task efficiently, as shown in Figure~\ref{fig:teaser}(a).

With sparse rewards, another problem is \textbf{forgetting}. Successful episodes may be rare, and subsequent policy updates can move the agent away from the successful behavior. Figure~\ref{fig:teaser}(b) illustrates this failure mode. In this case, the policy needs to rediscover the success through new interactions. The policy may then repeatedly forget and relearn the same successful behavior, leading to inefficient learning and unstable performance.

Existing approaches address parts of this problem through manually designed temporal preferences or experience replay. For distraction, per-step penalties can favor shorter trajectories~\citep{Szot2023SocialRearrangement,Liang2023ContextAware}, while dense-to-sparse reward schedules~\citep{luo2020balance} and time-limit handling~\citep{Pardo2018TimeLimits} alter the temporal structure of learning. These methods rely on hand-designed penalties, schedules, or horizons that require task-specific tuning. Weak temporal signals may have little effect, whereas strong ones may hinder exploration. They do not use the temporal structure of the agent's own successful behavior to set and continually refine targets for subsequent learning. For forgetting, standard self-imitation learning reuses high-return experience~\citep{Oh2018SIL}, which can help preserve success, but return alone may not distinguish fast successes from slower ones. This can focus the policy on irrelevant interactions instead of encouraging more efficient solutions.

\begin{tcolorbox}[
    colback=gray!8,
    colframe=black!20,
    boxrule=0.3pt,
    arc=1mm,
    left=0.5mm,
    right=0.5mm,
    top=0.3mm,
    bottom=0.3mm,
    before skip=6pt,
    after skip=6pt
]
\small
\textbf{Our central hypothesis}
is that temporally efficient successful behavior discovered during learning contains reusable self-supervisory signals for future policy improvement. Once a policy discovers that a task can be solved efficiently, subsequent learning should bias toward behaviors that match or exceed this efficiency, allowing temporal targets and efficient behaviors to be discovered and refined by the policy itself.
\end{tcolorbox}

We present \textbf{Temporal Self-Imitation Learning (TSIL)}, a reinforcement learning framework that mines temporal efficiency from self-generated successful trajectories and converts it into reusable supervision. TSIL uses fast successful trajectories in two complementary ways. First, the completion time of a successful trajectory becomes an adaptive temporal target that conditions future observations and rewards. As faster solutions emerge, this target tightens and encourages increasingly efficient behavior. Second, efficient successful trajectories are preserved in a replay buffer and reused through efficiency-weighted self-imitation, helping the policy revisit behaviors that may otherwise disappear under policy distribution shift.

We evaluate TSIL on 30 long-horizon tasks with dense and sparse rewards, state-based or image-based observations, and continuous or discrete control: 15 robot manipulation tasks spanning articulated-object interaction, insertion, tool use, transport, and contact-rich manipulation, and 15 MiniGrid tasks including navigation and object interaction challenges. Across these settings, TSIL consistently improves learning efficiency, behavioral efficiency, successful behavior revisitation, and robustness under unstable training conditions.

In summary, our contributions are:
\begin{itemize}[leftmargin=*,nosep]
    \item We identify temporal efficiency as a self-supervisory signal for long-horizon reinforcement learning, and show that reward alone can fail to distinguish efficient successful behavior from slower or reward-distracted success.
    \item We introduce Temporal Self-Imitation Learning, which combines adaptive temporal targets with efficiency-weighted self-imitation of successful trajectories.
    \item We demonstrate consistent improvements in learning efficiency, behavioral efficiency, successful behavior revisitation, and robustness across 30 long-horizon tasks spanning robot manipulation and MiniGrid.
\end{itemize}

\section{Related Work}
\label{sec:related_work}

\textbf{Temporal reinforcement learning.} Temporal reinforcement learning studies how time enters the decision process, from finite-horizon state augmentation and time-limit handling~\citep{Pardo2018TimeLimits} to continuous-time or discretization-robust control~\citep{Doya2000ContinuousTime,Tallec2019TimeDiscretization,Yildiz2021ContinuousTimeMBRL,Ramstedt2019RealTimeRL} and temporal abstraction through options~\citep{Sutton1999Options}, action repetition~\citep{Lakshminarayanan2017DAR,Sharma2017FiGAR}, or learned action timing~\citep{Biedenkapp2021TempoRL}. Recent time-aware RL methods further address irregular decision intervals~\citep{Kim2021TQN,Kim2023TimeAwareDRL}, time-step-conditioned world models~\citep{Nhu2025TimeAwareWorldModel}, and timing-aware driving or robot navigation~\citep{Li2024ActBetterTiming,Chen2023AFST}. Adjacent safe offline RL also conditions trajectory planning on dynamically specified deployment budgets, though these budgets denote cost constraints rather than task-time variables~\citep{Lin2023SafeOfflineBudget}. Closest to our setting, time-aware policy learning for robot control conditions a trained manipulation policy on remaining time and a time ratio, enabling inference-time tempo adaptation~\citep{Jia2025TimeAwarePolicy}. In contrast, TSIL uses time-awareness during training: the policy's own fast successes set configuration-conditioned adaptive temporal targets that reshape rewards, observations, and self-imitation priorities.

\textbf{Self-imitation learning.} Self-imitation learning uses an agent's own experience as supervision. The original SIL objective imitates high-return trajectories through advantage filtering~\citep{Oh2018SIL}, while later variants extend this idea with episodic hindsight replay~\citep{Dai2020EpisodicSIL}, sparse or delayed robot rewards~\citep{Chen2021SILCR}, and planning-based robot motion learning~\citep{Luo2021SILP,Luo2023SILPPlus}. In robot learning, successful experience has also been reused through sparse-reward relabeling~\citep{Bujalance2023RewardRelabelling}, task reduction~\citep{Li2021TaskReduction}, and goal-conditioned self-imitation~\citep{Li2022PAIR,Li2023GRSIL}. Recent self-improving robot systems further turn autonomous practice into supervision for manipulation and foundation-model post-training~\citep{Sharma2023SelfImprovingRobots,Bousmalis2023RoboCat,Ghasemipour2025SelfImprovingEFM}. TSIL differs by making replay selection temporal: it prioritizes fast successful trajectories and relabels them under the current adaptive temporal target.

\section{Temporal Self-Imitation Learning}
\label{sec:method}

The central idea of Temporal Self-Imitation Learning (TSIL) is to treat temporally efficient successful behavior during learning as reusable self-supervision for future policy improvement. Long-horizon reinforcement learning presents two coupled challenges. Policies can over-exploit dense rewards through slow or reward-distracted behavior rather than completing the task efficiently. Meanwhile, the rare fast successes that reveal better behavior may disappear during later on-policy updates due to distribution shift or unstable optimization. TSIL addresses these challenges by mining temporal efficiency from successful trajectories and feeding it back into future learning. The completion time of efficient successful trajectories becomes a progressively refined temporal target, while the trajectories themselves are replayed through efficiency-weighted self-imitation. Once efficient behavior emerges, future optimization increasingly biases toward reproducing and improving upon it.

\begin{figure}[t]
    \centering
    \includegraphics[width=\textwidth,trim=0 13bp 0 19bp,clip]{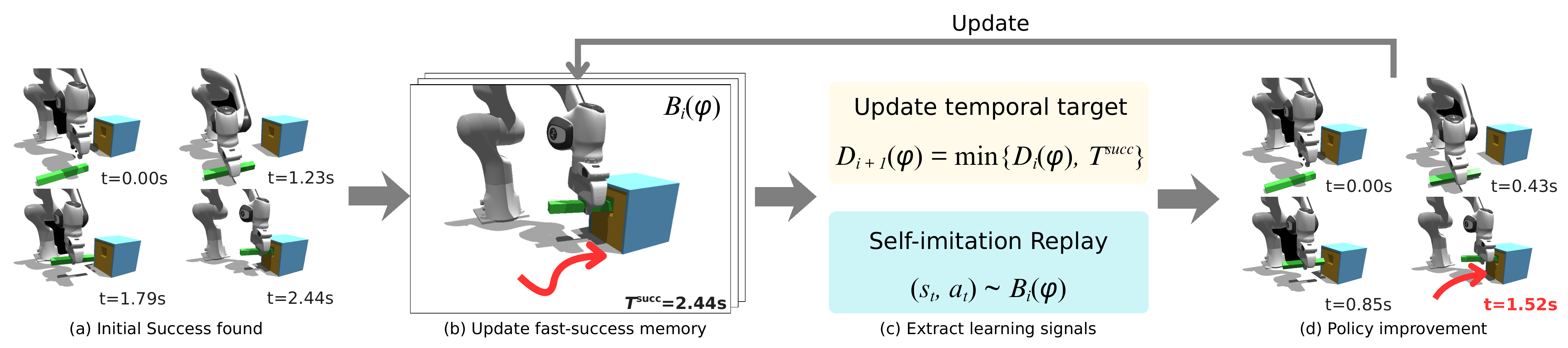}
    \caption{\textbf{TSIL training loop.} (a) The policy discovers a successful trajectory during early interactions. (b) The trajectory updates the policy's fast-success memory. (c) Its completion time updates the adaptive temporal target, while its transitions provide self-imitation supervision. (d) The updated policy is encouraged to complete the task more efficiently, and faster successful trajectories feed back into the memory.}
    \label{fig:tsil_method}
    \vspace{-1em}
\end{figure}

\subsection{Temporal target conditioning}
\label{sec:method_temporal_target}

Robot policies are commonly trained with an infinite-horizon discounted objective, $J(\pi)=\mathbb{E}_{\pi}[\sum_{t=0}^{\infty}\gamma^t r(s_t,a_t,s_{t+1})]$, using truncated rollouts and bootstrapping. We use an infinite-horizon formulation because the number of steps needed to complete a robot task is generally unknown. Task difficulty varies across configurations, and disturbances may require additional interaction for recovery. Although discounting weakly favors shorter trajectories, tuning it is a blunt temporal preference: small discount factors can suppress delayed success signals, while large ones only mildly favor faster completion. Moreover, it does not explicitly distinguish efficient successful behavior from slow or reward-distracted behavior. In long-horizon tasks, policies can still accumulate substantial return through prolonged recovery sequences or repeated exploitation of dense shaping rewards before eventually succeeding.

Instead of manually engineering temporal objectives, TSIL derives temporal supervision directly from successful behavior discovered during learning. Let $\phi$ denote a task configuration, such as initial state, goal, and environment configuration. Let $N$ be the episode horizon, $\Delta t$ the control interval, and $T^{\max}=N\Delta t$ the maximum rollout duration.

A central design choice is that temporal targets are configuration-conditioned rather than global. The same task can have very different feasible completion times depending on $\phi$, so a target discovered for an easy configuration should not be imposed on a harder one. For each training configuration, TSIL maintains an adaptive temporal target $D(\phi)\leq T^{\max}$, representing the fastest successful completion time discovered so far under comparable initial conditions. The policy receives the original observation together with elapsed interaction time and the active target, $\tilde{o}_t=(o_t,T_t^{\mathrm{used}},D(\phi))$, where $T_0^{\mathrm{used}}=0$ and $T_{t+1}^{\mathrm{used}}=T_t^{\mathrm{used}}+\Delta t$.

Importantly, the temporal target is not a hard environment constraint. Rollouts may continue beyond $D(\phi)$ until task completion or rollout truncation at $T^{\max}$. Instead, the temporal target acts as adaptive supervision that progressively biases learning toward increasingly efficient behavior. The resulting objective is $J(\pi;D)=\mathbb{E}_{\pi}[\sum_{t=0}^{\infty}\gamma^t r(s_t,a_t,s_{t+1},T_t^{\mathrm{used}},D(\phi))]$.

\subsection{Adaptive temporal supervision}
\label{sec:method_adaptive_target}

A successful trajectory reveals that a specific configuration can be solved within a particular amount of interaction time. Let $D_i$ denote the temporal target function at training iteration $i$, initialized as $D_1(\phi)=T^{\max}$. For a successful trajectory $\tau$, let $K_\tau$ be its completion step count and $T^{\mathrm{succ}}(\tau)=K_\tau\Delta t$. After each iteration, TSIL updates the target using the fastest successful trajectory discovered so far:
\begin{equation}
    D_{i+1}(\phi)
    =
    \min\!\left(
        D_i(\phi),
        \min_{\substack{\tau\in\mathcal{T}_i(\phi)\\ \mathbbm{1}_{\mathrm{succ}}(\tau)=1}}
        T^{\mathrm{succ}}(\tau)
    \right),
    \label{eq:addl}
\end{equation}
where the inner minimum is ignored if no successful trajectory for $\phi$ appears in $\mathcal{T}_i(\phi)$. Thus, temporal targets progressively tighten as the policy discovers more efficient successful behavior.

We further use the temporal target to shape the terminal success reward. For a trajectory terminating after $K$ steps with elapsed interaction time $T^{\mathrm{used}}=K\Delta t$, the target-conditioned success reward is
\begin{equation}
    r_{K-1}^{\mathrm{succ}}
    =
    r_{\mathrm{succ}}\,
    \mathbbm{1}_{\mathrm{succ}}
    \left[
        1+
        \min\!\left(
            \frac{D_i(\phi)}
            {T^{\mathrm{used}}},
            1
        \right)
    \right].
    \label{eq:target_reward}
\end{equation}
This reward preserves the task-completion incentive while adding a temporal bonus relative to the current target. Trajectories that finish before the target receive maximal temporal bonus, while slower trajectories receive less as $T^{\mathrm{used}}$ grows. With task rewards, the final reward is $r_t=r_t^{\mathrm{task}}+r_t^{\mathrm{succ}}$.

\subsection{Efficiency-weighted self-imitation}
\label{sec:method_sil}

Adaptive temporal supervision biases the current rollout toward efficient behavior, but efficient successful trajectories may still disappear during later policy updates. TSIL therefore stores temporally efficient successful trajectories in an auxiliary replay buffer. Unlike standard self-imitation learning, which prioritizes high-return trajectories~\citep{Oh2018SIL}, TSIL prioritizes fast successful trajectories because high return can still contain slow or reward-distracted behavior.

For each configuration $\phi$, TSIL maintains a replay buffer $\mathcal{B}_i(\phi)$ of capacity $k$ that retains the fastest successful trajectories discovered so far. Before replay, the current replay buffer is relabeled using the current target $D_i$, yielding $\hat{\mathcal{B}}_i(\phi)$. For each trajectory $\tau$ collected under configuration $\phi^\tau$, the observation becomes $\hat{o}_t^\tau=(o_t^\tau,T_t^{\mathrm{used},\tau},D_i(\phi^\tau))$, the reward becomes $\hat{r}_t^\tau=r_t^{\mathrm{task},\tau}+\mathbbm{1}(t=K_\tau-1)\hat{r}_{K_\tau-1}^{\mathrm{succ},\tau}$, and the relabeled return is $\hat{G}_t^\tau=\sum_{j=t}^{|\tau|-1}\gamma^{j-t}\hat{r}_j^\tau$, where $|\tau|$ is the stored trajectory length.

Not every transition in a successful trajectory should be imitated. Following self-imitation learning, TSIL uses a positive return-value gap, $A_t^+(\tau)=[\hat{G}_t^\tau-V_\psi(\hat{o}_t^\tau)]_+$, as a soft gate so replay pressure vanishes once the critic judges the current policy to match the stored behavior. To further prioritize temporally efficient success, trajectories are weighted by $w_i(\tau)=1+\mathbbm{1}_{\mathrm{succ}}(\tau)\min(D_i(\phi^\tau)/T^{\mathrm{succ}}(\tau),1)$.

The full objective is $\mathcal{L}=\mathcal{L}_{\mathrm{PPO}}+\lambda_{\mathrm{SI}}\mathcal{L}_{\mathrm{SI}}$, where the auxiliary self-imitation loss is
\begin{equation}
    \mathcal{L}_{\mathrm{SI}}
    =
    \mathbb{E}_{(\tau,t)\sim\hat{\mathcal{B}}_i}
    \left[
        -w_i(\tau)A_t^+(\tau)\log\pi_\theta(a_t^\tau\mid\hat{o}_t^\tau)
        +
        \frac{\lambda_{\mathrm{V,SI}}}{2}
        \left(A_t^+(\tau)\right)^2
    \right].
    \label{eq:sil_loss}
\end{equation}

\subsection{Implementation details}
\label{sec:implementation}

Algorithm~\ref{alg:tasil} in Appendix~\ref{app:optimization} summarizes the complete TSIL training procedure. TSIL was implemented on top of PPO using a standard actor-critic architecture. PPO itself is unchanged except for temporal target conditioning, target-conditioned rewards, and the auxiliary efficiency-weighted self-imitation objective. Training uses parallel simulation environments, each maintaining its own task configuration, temporal target, and replay buffer. To balance replay across environments and prevent easy configurations from dominating it, each configuration stores the top-$k$ ($k=5$) fastest successful trajectories. During very early training, when successful trajectories are scarce, the replay buffer is supplemented with high-return trajectories to stabilize replay. Appendix~\ref{app:optimization} describes fallback replay and bootstrapped returns.

Within iteration $i$, rollout collection, replay relabeling, and policy optimization all use $D_i$. The updated target $D_{i+1}$ is committed after optimization and applied at the next episode reset, keeping the rollout observations, rewards, and stored old-policy log probabilities consistent during the optimization. See Appendix~\ref{app:optimization} for the update schedule.

\textbf{Replay sampling.} TSIL filters transitions using the positive return-value gap and samples up to $n$ positive transitions per trajectory without replacement, with probability proportional to the gap. The replay set is constructed once per PPO update, contains at most $k n N_{\mathrm{env}}$ transitions, and is reused across $E$ PPO epochs. Sampling details are provided in Appendix~\ref{app:method_details}.

\textbf{Evaluation temporal target inference.} For an unseen MiniGrid evaluation configuration, TSIL sets the temporal target to the mean target of up to five nearest successful (5-NN) training configurations from the same task. If no successful training configuration is available, TSIL uses the episode horizon. The inferred target remains fixed during evaluation. Configuration representations and distance measures are specified in Appendix~\ref{app:method_details}.

\Needspace{5\baselineskip}
\section{Experiments}
\label{sec:experiments}

\begin{figure}[t]
    \centering
    \includegraphics[width=0.85\linewidth,trim=0 3bp 0 8bp,clip]{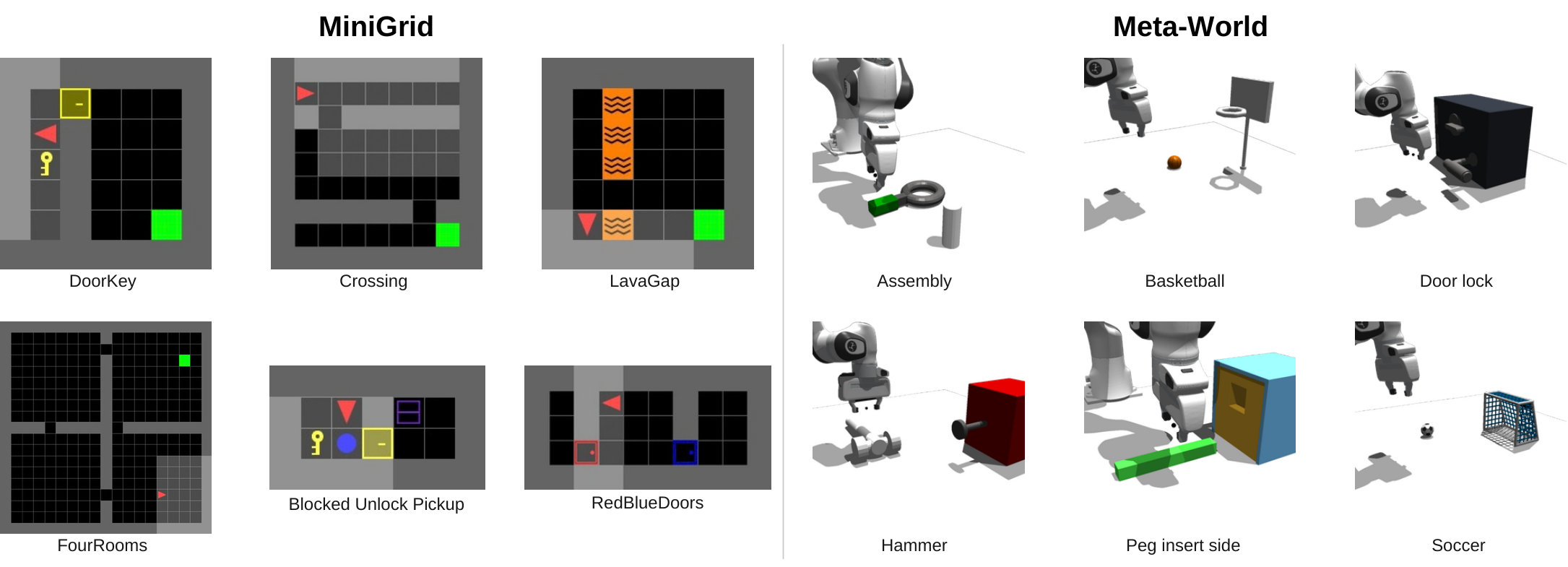}

    \caption{\textbf{Representative tasks.} Left: MiniGrid tasks cover navigation and object interaction with sparse rewards, image observations, and discrete control. Right: Meta-World tasks cover assembly, tool use, articulated object interaction, and dynamic manipulation with dense rewards, state observations, and continuous control.}
    \label{fig:task_contact_sheet}
    \vspace{-1.0em}

\end{figure}

We evaluate TSIL by asking three questions: \textbf{(1) Effectiveness:} does TSIL improve performance on long-horizon reinforcement learning tasks? \textbf{(2) Efficiency:} does TSIL improve both learning efficiency and behavioral efficiency? \textbf{(3) Robustness:} does TSIL improve robustness under unstable training conditions?

\textbf{Tasks and training settings.} We evaluate TSIL on 30 long-horizon tasks from two complementary domains. The 15 MiniGrid tasks~\citep{MinigridMiniworld23} use partial image observations and discrete control with sparse rewards, spanning navigation and object-interaction challenges. The 15 Meta-World manipulation tasks~\citep{yu2020meta}, implemented in Isaac Gym through MTBench~\citep{joshi2025benchmarking}, use state observations and continuous robot control with dense shaping rewards. Each task is trained independently under its environment-specific horizon and interaction budget. For robustness tests, we apply policy-gradient noise and increased PPO clip ratios to both benchmarks, plus dense-reward dropout to Meta-World because MiniGrid rewards are already sparse. Implementation and evaluation details are provided in Appendix~\ref{app:implementation}.

\textbf{Baselines.} All methods use PPO~\citep{Schulman2017PPO} as the base optimizer. We compare against (1) Vanilla PPO, (2) Step-cost~\citep{Szot2023SocialRearrangement,Liang2023ContextAware}, which subtracts the per-step cost $c(\phi)=0.1r_{\mathrm{succ}}/T^{\max}(\phi)$ and avoids excessive cumulative penalties for long-horizon tasks, (3) dense-to-sparse reward scheduling (D2S)~\citep{luo2020balance}, which removes dense rewards at 50\% of the training budget, (4) standard self-imitation learning (SIL)~\citep{Oh2018SIL}, (5) fixed temporal target learning (FTTL), which observes elapsed time and a fixed target equal to the episode horizon, and (6) adaptive temporal target learning without fast-success replay (ATTL). For fair comparison, SIL matches TSIL's replay budget, transition sampler, and optimization schedule, but stores the top-$k$ ($k=5$) highest-return trajectories per configuration instead of fast successes. It also disables adaptive temporal-target conditioning (elapsed-time and target observations) and temporal-efficiency weighting. All methods use the same base success-reward coefficient $r_{\mathrm{succ}}$ within each benchmark. Baseline settings and hyperparameters are listed in Appendix~\ref{app:baselines}.

\textbf{Metrics.} We report five primary metrics. (1) Training success rate is the best rolling success rate within the final 10\% of training. (2) Area under the success rate curve (AUC) summarizes learning performance over the full interaction budget. (3) Successful episode count is the total number of successful episodes completed during training. (4) Normalized completion time is $T^{\text{succ}}/T^{\max}$ on success and $1$ on failure, with lower values indicating more reliable and efficient completion. This is evaluated at the checkpoint with the best training success rate. (5) Evaluation success rate measures success over 2,000 evaluation episodes per task per seed, using the checkpoint with the highest rolling training success rate within the final 10\% of training. This tests generalization to unseen configurations, including inference of their temporal targets. To study how adaptive temporal targets and efficiency-weighted replay alter learning, we additionally analyze positive-advantage mass, gradient update directions, and an efficiency-weighting ablation. Each method is evaluated with three random seeds per task. More details are provided in Appendix~\ref{app:metrics}.

\subsection{TSIL improves effectiveness, learning and behavioral efficiency}
\label{sec:exp_main}

\begin{figure}[t]
    \centering
    \includegraphics[width=0.95\textwidth]{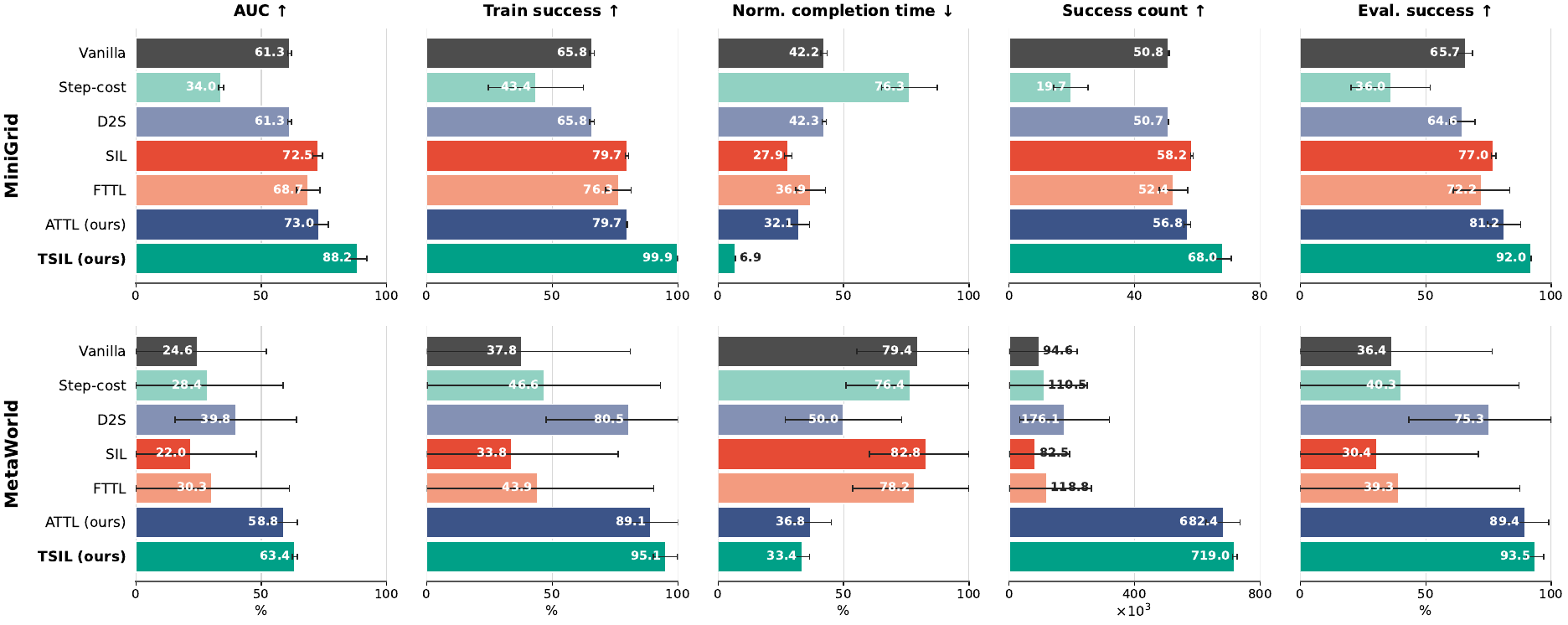}
    \caption{\textbf{Main results under standard training}. TSIL performs best overall in task success, learning efficiency, and behavioral efficiency. Values are mean $\pm$ standard deviation across tasks and seeds.}
    \label{fig:main_results}
    \vspace{-1.0em}
\end{figure}

As shown in Fig.~\ref{fig:teaser}(c) and Fig.~\ref{fig:main_results}, TSIL achieved the strongest overall performance across effectiveness, learning efficiency, and behavioral efficiency in both Meta-World and MiniGrid. In particular, TSIL achieved the highest training success rate and AUC while also obtaining the lowest normalized completion time, the largest number of successful training episodes, and the highest evaluation success rate. These results support our key hypothesis that temporally efficient successful trajectories provide self-supervision for policy improvement. Appendix~\ref{app:additional_results} provides success-rate and completion-time training curves for all 15 tasks in each benchmark.

The progression across baselines further clarifies the contribution of each component. Reward-engineering baselines (Step-cost and D2S) improved some individual metrics, but remained substantially weaker than temporal-target-based methods in overall success and efficiency. FTTL improved training success over Vanilla in both domains, showing that temporal conditioning alone was beneficial, yet its fixed temporal target did not consistently encourage successful and efficient task completion. Introducing adaptive temporal targets (ATTL) substantially improved learning and behavioral efficiency by progressively tightening supervision toward the shorter temporal targets discovered during learning. This reduced distraction from slow or suboptimal behavior and improved both training and evaluation success rates. SIL improved success and learning efficiency in MiniGrid but remained weaker on Meta-World. This suggests that broader high-return experience can help in the sparse-reward environment, but high-return replay can also direct learning away from efficient task completion in dense-reward settings. In contrast, full TSIL consistently achieved the best overall performance in task success, learning efficiency, and behavioral efficiency, demonstrating that replaying temporally efficient successful trajectories provided substantially more effective supervision than replaying generic high-return experience. The high evaluation success rate also demonstrates strong generalization of the estimated temporal targets to new configurations. We also observed fewer unnecessary motions, parallel execution of motions, and more effective use of momentum within the controller\textquotesingle s joint-velocity limits in Meta-World tasks. See the supplementary video for examples.

\subsection{TSIL improves robustness under unstable training}
\label{sec:exp_stability}

Long-horizon reinforcement learning is often highly sensitive to optimization instability. In practice, unreliable policy gradients and imperfect reward signals can cause policies to drift away from previously discovered efficient successful behaviors. Since TSIL explicitly preserves and replays temporally efficient successful trajectories, we hypothesize that it should also improve robustness under unstable training conditions by anchoring optimization to efficient successful behavior even when the current on-policy signal becomes unreliable. To evaluate this, we introduced three training disturbances for Meta-World and the two applicable disturbances for MiniGrid as discussed above.

\begin{figure}[!t]
    \centering
    \includegraphics[width=\linewidth]{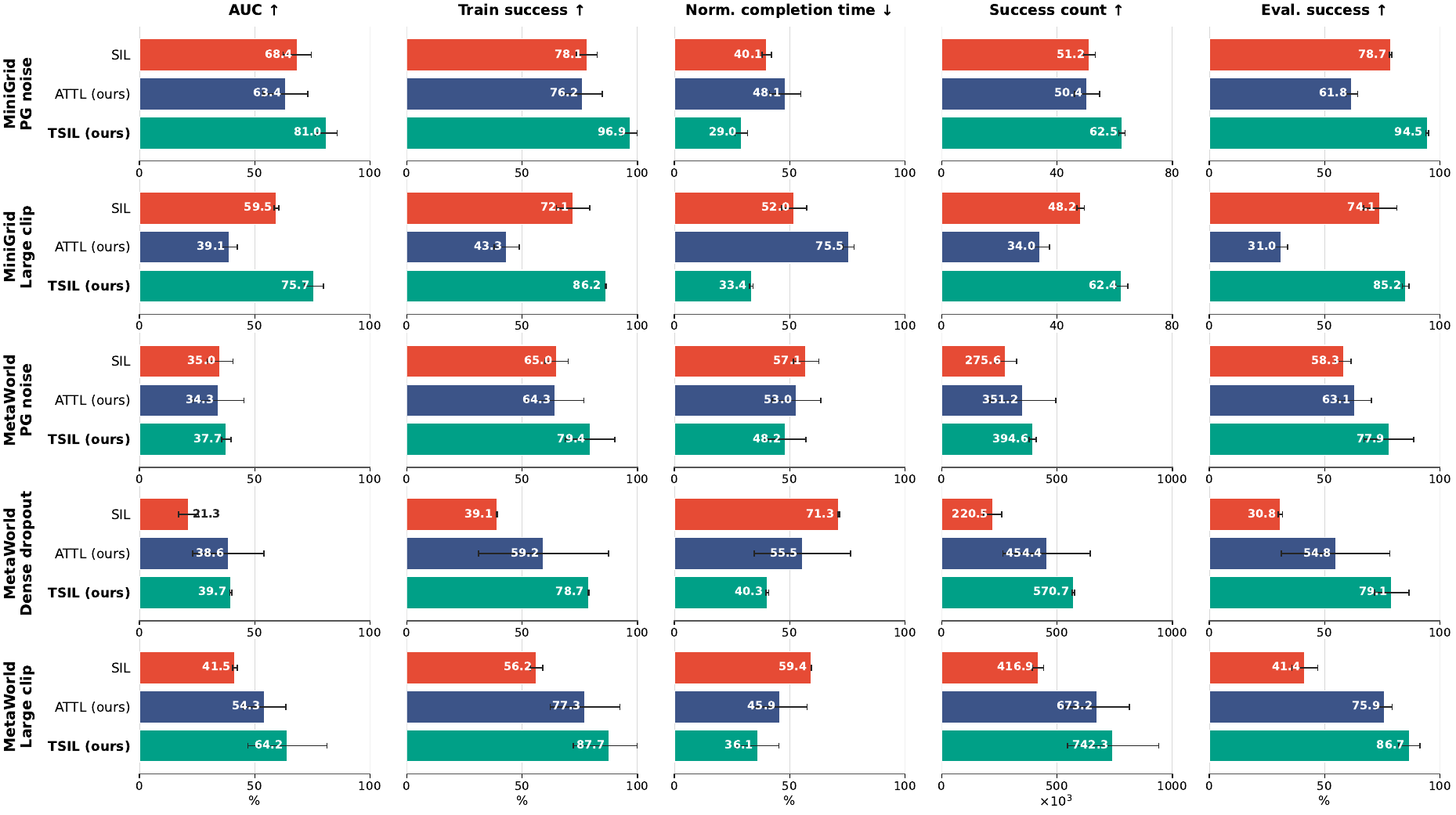}
    \caption{\footnotesize\textbf{Main results under disturbed training.} ATTL degrades under disturbances. SIL improves MiniGrid but remains weaker in Meta-World. TSIL performs best overall, especially under severe policy drift and unreliable dense rewards. Values are mean $\pm$ standard deviation across disturbance levels, tasks, and seeds.}
    \label{fig:stability_5metrics}
    \vspace{-1.0em}
\end{figure}

Figure~\ref{fig:stability_5metrics} reports the same five metrics for each disturbance class (settings in Appendix~\ref{app:disturbance_settings}). ATTL shows large performance drops across all disturbances. Without fast-success replay, unstable policy updates can rapidly move the policy away from previously discovered successful behaviors and lead to policy collapse. When dense reward supervision is unstable, ATTL may forget previously learned behavior and repeatedly rediscover it, leading to degraded performance and slower learning. TSIL consistently achieved the strongest overall performance, indicating that replaying efficient successful trajectories provided an effective optimization anchor against policy drift, behavior forgetting, and unreliable reward signals.

The disturbances also distinguish generic high-return replay from temporally efficient replay. SIL maintained competitive performance on MiniGrid and slightly exceeded ATTL on several metrics, but performed even worse in Meta-World. This comparison exposes a limitation of high-return-based replay: in dense-reward tasks, high return may correspond to slow, failed, or reward-distracted behavior and becomes even less reliable under reward dropout. Replay can mitigate forgetting, but its selection criterion determines whether it preserves behavior that supports efficient task completion. In contrast, TSIL selects replay trajectories by successful completion and temporal efficiency. TSIL's clearer gains across all Meta-World disturbances indicate that fast-success replay anchors learning to efficient and reliable behavior, providing both effective supervision and stabilizing memory signals under unreliable rewards and unstable optimization.

\Needspace{0.40\textheight}
\subsection{Analysis}
\label{sec:exp_analysis}

\begin{wrapfigure}{r}{0.65\linewidth}
    \centering
    \vspace{-2.7em}
    \includegraphics[width=\linewidth,trim=0 3bp 0 18bp,clip]{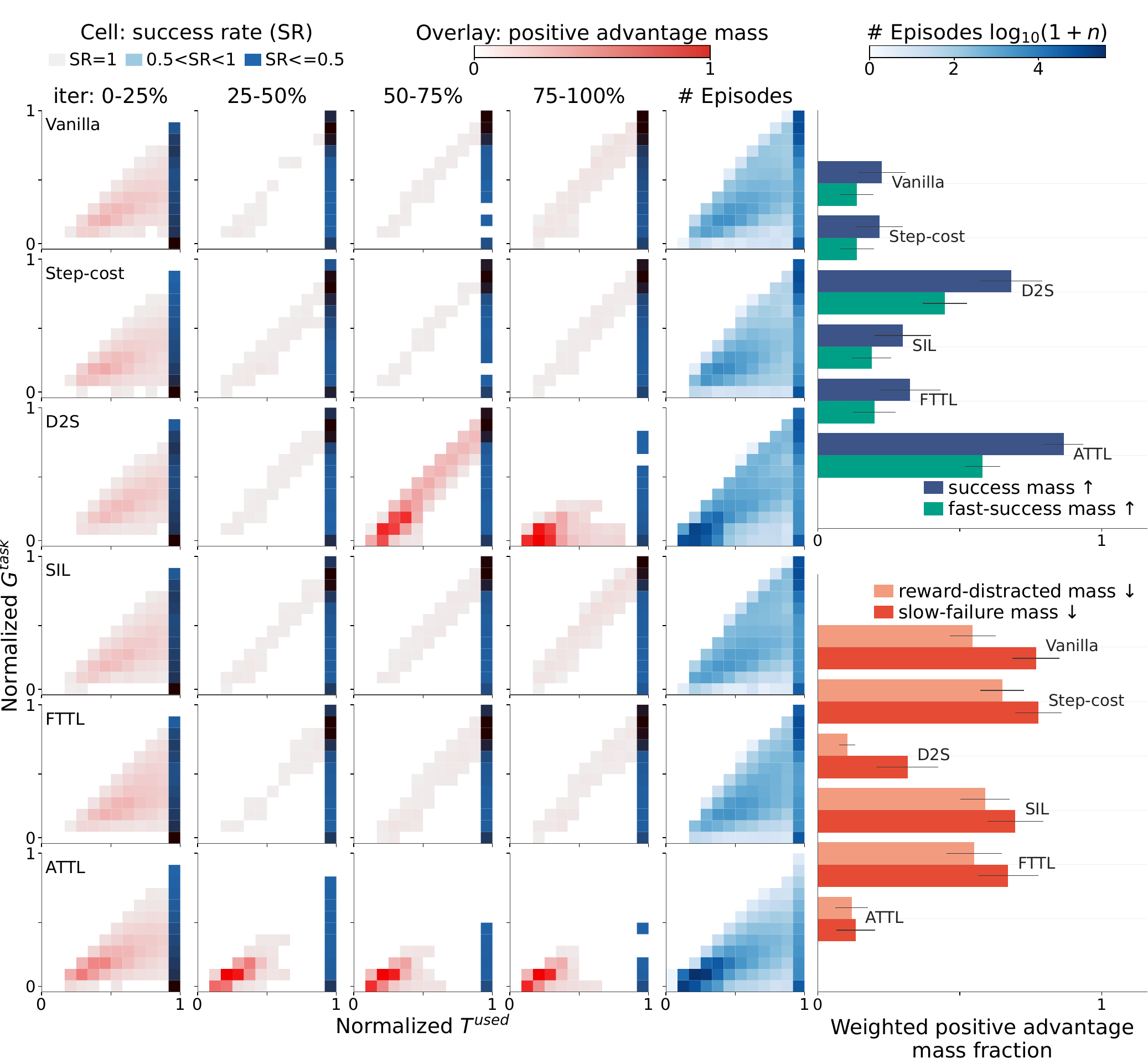}
    \caption{\footnotesize\textbf{Adaptive temporal targets redirect PPO optimization.} Adaptive temporal targets rapidly concentrate positive PPO update signal on fast successful trajectories while suppressing reward-distracted and slow-failure behavior. Left: one representative Peg Insertion Side learning signal map. Right: aggregated results over all 15 tasks.}
    \vspace{-1.7em}
    \label{fig:addl_signal}
\end{wrapfigure}

\leavevmode
\textbf{Why adaptive temporal targets help.} We next analyzed how adaptive temporal targets altered PPO optimization before replay was added using Meta-World tasks. For each episode, we aggregated the positive-advantage mass used by PPO and grouped episodes by normalized completion time and dense task return. In Fig.~\ref{fig:addl_signal}, cell color indicates empirical success rate, while the overlaid color gradient indicates the fraction of positive-advantage mass assigned to that region. The summary bars further aggregate this update signal into four categories: success, fast-success, slow-failure, and reward-distracted behavior. Definitions and additional tasks: Appendices~\ref{app:diagnostic_metrics}--\ref{app:learning_signals}.

As shown in Fig.~\ref{fig:addl_signal}, adaptive temporal targets rapidly concentrated PPO update signal on fast successful trajectories while \mbox{assigning} little optimization pressure to reward-distracted or slow-failure behavior. Fixed temporal targets improved time awareness but could not distinguish achievable efficient behavior from generic timing constraints, allowing failed high-return trajectories to continue receiving positive updates. Vanilla, Step-cost, and SIL were substantially more vulnerable to reward distraction. D2S partially mitigated this effect by removing dense rewards at the cost of losing shaped guidance and wasting much of the training budget before switching. These results suggest that adaptive temporal targets improve behavioral efficiency by progressively redirecting reinforcement pressure toward increasingly efficient successful behavior discovered during learning.

\begin{figure}[!t]
    \centering
    \includegraphics[width=0.95\linewidth,trim=1bp 4bp 5bp 14bp,clip]{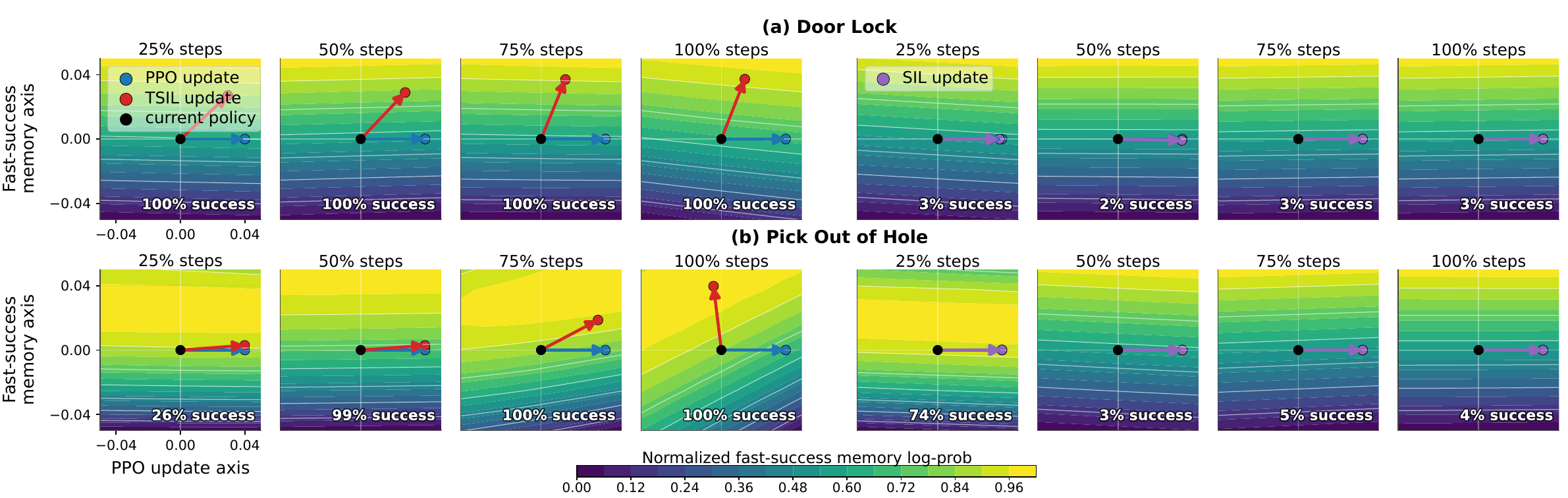}
    \caption{\footnotesize\textbf{TSIL redirects updates toward fast-success behavior.} Across four training stages, TSIL aligns updates more strongly than SIL with high fast-success-memory likelihood. In (a), this promotes rapid exploitation of early successes. In (b), it preserves successful behavior and reduces later forgetting.}
    \label{fig:replay_landscape}
    \vspace{-1.0em}
\end{figure}

\Needspace{0.24\textheight}
\begin{wraptable}{r}{0.50\linewidth}
    \centering
    \vspace{-1.0em}
    \caption{\textbf{Efficiency-weighting (EW) ablation in Meta-World.} Without efficiency weighting, TSIL shows lower performance, reported as mean $\pm$ SD across tasks and seeds.}
    \label{tab:efficiency_weighting_ablation}
    \scriptsize
    \setlength{\tabcolsep}{1.5pt}
    \begin{tabular}{lccc}
        \toprule
        Method & AUC (\%) $\uparrow$ & Success (\%) $\uparrow$ & Norm. time (\%) $\downarrow$ \\
        \midrule
        TSIL (no EW) & $57.2 \pm 6.6$ & $89.31 \pm 5.04$ & $38.6 \pm 4.7$ \\
        TSIL & $\mathbf{63.4 \pm 1.0}$ & $\mathbf{95.1 \pm 4.8}$ & $\mathbf{33.4 \pm 3.1}$ \\
        \bottomrule
    \end{tabular}
    \vspace{-1.0em}
\end{wraptable}

\leavevmode
\textbf{Why efficiency-weighted replay helps.} Adaptive temporal targets bias the current PPO update toward efficient successful behavior, but later on-policy updates can still move the policy away from those trajectories. TSIL addresses this by preserving temporally efficient successful trajectories in replay and explicitly increasing their likelihood during optimization. Figure~\ref{fig:replay_landscape} further shows that TSIL redirects PPO updates toward regions with high fast-success-memory likelihood and maintains successful behavior, whereas generic SIL provides weaker alignment and does not reliably preserve success. Landscape construction and additional tasks are provided in Appendices~\ref{app:landscape_method}--\ref{app:policy_landscapes}. Table~\ref{tab:efficiency_weighting_ablation} compares TSIL with TSIL (no efficiency-weighting) ablation, isolating the contribution of efficiency weighting while holding fast-success replay fixed. Removing the efficiency weighting lowers AUC and success rate and increases normalized completion time, showing that prioritizing more efficient successes improves both learning and behavioral efficiency.


\section{Conclusion \& Limitations}
\label{sec:conclusion}

We introduced Temporal Self-Imitation Learning (TSIL), which uses temporally efficient successful trajectories as reusable self-supervision through adaptive temporal targets and efficiency-weighted replay. Across 30 long-horizon manipulation and navigation tasks, including disturbed training settings, TSIL improves task success, learning efficiency, behavioral efficiency, and robustness. Our analysis shows that adaptive targets focus learning on efficient success, while replay helps preserve behaviors that unstable on-policy updates might otherwise erase.

\textbf{Limitations.} TSIL does not directly solve initial exploration. It improves exploitation and consolidation after rare successful trajectories appear, so the base learner still needs enough exploration or task guidance to discover success. Temporal efficiency is also only one behavioral signal considered here. If deployment requires diverse modes such as slower, safer, smoother, or more stable behavior, future systems could first learn reliable efficient behavior and then combine temporal targets with additional metrics that reflect those requirements. Finally, TSIL requires storing successful trajectories in memory. We used a small fixed-size replay buffer, whereas more memory-efficient replay and compression could preserve fast successes more effectively.

\clearpage
\section*{AI use statement}
We used an LLM-based tool to polish writing, identify additional relevant literature, and inspect code. The authors reviewed all AI-assisted output and take responsibility for the final manuscript.

\ificlrfinal
\section*{Acknowledgments}
This work is supported by DARPA TIAMAT program under award HR00112490419, ARO under award W911NF2410405, NSF ERC PreMiEr under award 2133504, and ARL STRONG program under awards W911NF2320182, W911NF2220113, and W911NF242021.
\fi

\bibliographystyle{iclr2027_conference}
\bibliography{reference}

\clearpage
\appendix
\section{Algorithm and Implementation Details}
\label{app:optimization}

\subsection{Algorithm details}

\textbf{Optimization objective.}
TSIL uses the same PPO objective as all baselines; temporal target conditioning only changes the observation and reward passed to PPO.
For rollout data $\mathcal{T}_i$ collected at training iteration $i$, define the PPO probability ratio
\begin{equation}
    \rho_t(\theta)
    =
    \frac{\pi_\theta(a_t\mid\tilde{o}_t)}
    {\pi_{\theta_{\mathrm{old}}}(a_t\mid\tilde{o}_t)}.
    \label{eq:appendix_ppo_ratio}
\end{equation}
\begin{equation}
    \bar{\rho}_t(\theta)
    =
    \operatorname{clip}(\rho_t(\theta),1-\epsilon,1+\epsilon).
    \label{eq:appendix_ppo_clip}
\end{equation}
\begin{equation}
    \ell_t^{\mathrm{clip}}(\theta)
    =
    -\min\!\left(\rho_t(\theta)A_t,\bar{\rho}_t(\theta)A_t\right).
    \label{eq:appendix_ppo_surrogate}
\end{equation}
\begin{equation}
    \mathcal{L}_{\mathrm{PPO}}
    =
    \mathbb{E}_{t\sim\mathcal{T}_i}
    \left[
        \ell_t^{\mathrm{clip}}(\theta)
        +
        \frac{c_V}{2}\left(V_\psi(\tilde{o}_t)-G_t\right)^2
        -
        c_H\mathcal{H}\!\left(\pi_\theta(\cdot\mid\tilde{o}_t)\right)
    \right].
    \label{eq:appendix_ppo_loss}
\end{equation}
\begin{equation}
    \mathcal{L}
    =
    \mathcal{L}_{\mathrm{PPO}}
    +
    \lambda_{\mathrm{SI}}\mathcal{L}_{\mathrm{SI}}.
    \label{eq:appendix_full_loss}
\end{equation}
Here $\pi_\theta$ is the current actor with parameters $\theta$, $\pi_{\theta_{\mathrm{old}}}$ is the rollout policy used to collect $\mathcal{T}_i$, $a_t$ is the action sampled at time step $t$, and $\tilde{o}_t$ is the temporally augmented observation.
The operator $\operatorname{clip}(x,l,u)=\min(\max(x,l),u)$ truncates its first argument to the interval $[l,u]$, $\epsilon$ is the PPO clipping threshold, $\bar{\rho}_t$ is the clipped probability ratio, and $\ell_t^{\mathrm{clip}}$ is the per-transition clipped policy loss.
$A_t$ and $G_t$ are the on-policy PPO advantage and return estimates, $V_\psi$ is the critic with parameters $\psi$, $c_V$ and $c_H$ are the value-loss and entropy coefficients, and $\mathcal{H}(\pi_\theta(\cdot\mid\tilde{o}_t))$ is the policy entropy.
The expectation $\mathbb{E}_{t\sim\mathcal{T}_i}$ averages over rollout timesteps in $\mathcal{T}_i$.
The self-imitation coefficient is $\lambda_{\mathrm{SI}}$, and the self-imitation loss $\mathcal{L}_{\mathrm{SI}}$ is defined in Eq.~\eqref{eq:sil_loss}.

\begin{algorithm}[b!]
\caption{Temporal Self-Imitation Learning}
\label{alg:tasil}
\footnotesize
\begin{algorithmic}[1]
\REQUIRE configurations $\{\phi_m\}_{m=1}^M$, horizon $N$, interval $\Delta t$, replay buffer size $k$, PPO epochs $E$, step size $\eta$, SIL coefficients $\lambda_{\mathrm{SI}},\lambda_{\mathrm{V,SI}}$
\STATE $\theta,\psi\leftarrow$ initialize actor and critic; $\mathcal{B}_0(\phi)\leftarrow\emptyset$; $D_1(\phi)\leftarrow T^{\max}$
\FOR{iteration $i=1,2,\ldots$}
    \STATE $\mathcal{T}_i\leftarrow\operatorname{Rollout}(\pi_\theta,D_i)$, with $\tilde{o}_t=(o_t,T^{\mathrm{used}}_t,D_i(\phi))$, $r_t=r_t^{\mathrm{task}}+r_t^{\mathrm{succ}}$ \hfill Eq.~\eqref{eq:target_reward}
    \STATE $\mathcal{B}_{i}(\phi)\leftarrow\operatorname{TopKFastSuccess}(\mathcal{B}_{i-1}(\phi)\cup\mathcal{T}_i(\phi),k)$ \hfill Eq.~\eqref{eq:appendix_topk_fast}
    \STATE $\hat{\mathcal{B}}_{i}(\phi)\leftarrow\operatorname{Relabel}(\mathcal{B}_{i}(\phi),D_i)$; compute $\hat{G}^{\tau}_t$, $A^+_t(\tau)$, $w_i(\tau)$ \hfill Sec.~\ref{sec:method_sil}, Eq.~\eqref{eq:appendix_bootstrap_return}
    \FOR{$e=1,\ldots,E$}
        \STATE $\mathcal{L}\leftarrow\mathcal{L}_{\mathrm{PPO}}(\theta,\psi;\mathcal{T}_i)+\lambda_{\mathrm{SI}}\mathcal{L}_{\mathrm{SI}}(\theta,\psi;\hat{\mathcal{B}}_{i})$ \hfill Eq.~\eqref{eq:appendix_full_loss}
        \STATE $(\theta,\psi)\leftarrow(\theta,\psi)-\eta\nabla_{\theta,\psi}\mathcal{L}$
    \ENDFOR
    \STATE $D_{i+1}\leftarrow\operatorname{TargetUpdate}(D_i,\mathcal{T}_i)$ \hfill Eq.~\eqref{eq:addl}
\ENDFOR
\end{algorithmic}
\end{algorithm}

\textbf{Replay buffer update.}
The replay buffer is intended to preserve behaviors that show a configuration can be solved efficiently.
For each configuration, TSIL therefore keeps at most $k$ successful trajectories with the shortest completion times and discards slower successful trajectories once the replay buffer is full.
Given a candidate set $\mathcal{S}$, this update is
\begin{equation}
    \operatorname{TopKFastSuccess}(\mathcal{S},k)
    =
    \operatorname{sort}_{T^{\mathrm{succ}}\uparrow}
    \left(
        \{\tau\in\mathcal{S}\mid \mathbbm{1}_{\mathrm{succ}}(\tau)=1\}
    \right)_{1:k}.
    \label{eq:appendix_topk_fast}
\end{equation}
where $T^{\mathrm{succ}}(\tau)$ is the completion time of trajectory $\tau$.
The arrow $T^{\mathrm{succ}}\uparrow$ denotes ascending order, so smaller completion times are ranked first and retained with higher priority.

\textbf{Fallback replay.}
For compactness, Algorithm~\ref{alg:tasil} shows the fast-success replay buffer update. In implementation, the replay buffer also maintains a return-ranked fallback pool.
When fewer than $k$ successful trajectories are available for the fast-success replay buffer of a configuration, the remaining replay slots are filled by top-return trajectories from the same configuration to stabilize early minibatch updates.
These fallback trajectories use the same target-conditioned relabeling as fast successes.
Let $|\tau|$ denote the stored length of trajectory $\tau$ in replay transitions.
Relabeled returns include a value bootstrap for truncated trajectories,
\begin{equation}
    \hat{G}^{\tau}_t
    =
    \sum_{j=t}^{|\tau|-1}\gamma^{j-t}\hat{r}^{\tau}_j
    +
    \gamma^{|\tau|-t}(1-d_\tau)V_\psi(\hat{o}^{\tau}_{|\tau|}),
    \label{eq:appendix_bootstrap_return}
\end{equation}
Here $d_\tau=1$ for terminal success or failure, and $d_\tau=0$ for truncation without success; only the latter receives a value bootstrap.

\textbf{On-policy training clarification.}
The PPO loss is computed only from the freshly collected on-policy rollout batch $\mathcal{T}_i$.
Stored trajectories do not enter the PPO likelihood ratio or an off-policy Bellman backup.
They are used only through the auxiliary self-imitation loss in Eq.~\eqref{eq:sil_loss}.
The value term in $\mathcal{L}_{\mathrm{SI}}$ regresses the critic toward relabeled Monte Carlo or bootstrapped replay-buffer returns when the positive return-value gap is nonzero. This is auxiliary value supervision rather than a one-step off-policy Bellman target.
Thus, TSIL remains an on-policy PPO method augmented with a small fast-success memory for self-imitation.

\subsection{Implementation details}
\label{app:method_details}

\textbf{Parallel-environment implementation.}
Temporal efficiency is configuration-conditioned: a short completion time for a simple configuration may not be enough for a hard configuration.
We therefore fix the configuration within each parallel environment: each environment keeps a persistent configuration $\phi$ across resets, while configurations are randomly sampled independently across environments.
TSIL maintains the adaptive temporal target $D(\phi)$ and replay buffer independently for each parallel environment.

\textbf{Unusually fast successes.} Based on our parallel-environments setting, an unusually fast success tightens only the temporal target of its own configuration, leaving the targets of other configurations unchanged. Its trajectory is retained among the fastest successes and replayed to encourage the policy to reproduce the behavior. Moreover, since the target remains soft, even slower successes still receive the base success reward plus the temporal bonus in Eq.~\eqref{eq:target_reward}. For successful completion times above the target, this bonus increases as completion time approaches the target, so matching the exact fastest observed time is not required to receive a learning signal. These mechanisms preserve supervision even when the fastest success is difficult to repeat.

\textbf{Replay relabeling.}
During replay, stored trajectories keep their original observations, actions, and elapsed-time channel, but the target channel is relabeled using the current temporal target $D(\phi)$.
The task reward is unchanged, and the terminal temporal success reward is recomputed under the current target.
Replay returns use the relabeled rewards; truncated fallback trajectories additionally use the value bootstrap in Eq.~\eqref{eq:appendix_bootstrap_return}.

\textbf{Replay sampling.}
For each configuration, TSIL selects up to $k$ stored trajectories, prioritizing the fastest successes and filling unused slots with the highest-return trajectories. TSIL screens all valid transitions using the positive return-value gap and samples up to $n$ positive transitions per trajectory without replacement, with probability proportional to the gap. The replay set is constructed once per PPO update and reused across $E$ epochs. With $N_{\mathrm{env}}$ parallel environments, it contains at most $k n N_{\mathrm{env}}$ transitions and can be smaller when fewer positive transitions are available at early training.

\textbf{Evaluation temporal target inference.} During the evaluation stage for unseen configurations, TSIL uses the mean temporal target of up to five nearest successful training configurations from the same task. If none of the task's training configurations has succeeded, TSIL uses the maximum episode horizon. The inferred target remains fixed throughout evaluation. Configuration representations are provided by the task environment described below.

\textbf{MiniGrid configuration.} The configuration $\phi$ includes the initial symbolic grid, agent position and orientation, and mission where applicable. Within each task, distance combines the fraction of grid cells with different object/color/state encodings, the agent\textquotesingle s Manhattan displacement normalized by grid dimensions, and penalties for orientation and mission mismatches.

\textbf{Meta-World configuration.} The configuration $\phi$ is represented by the task\textquotesingle s object/goal randomization vector. Within each task, distance is Euclidean distance between these vectors after scaling each feature by its range across training configurations.

\textbf{Normalization.}
All methods use the same normalization pipeline.
Physical observation and state features use running mean--standard-deviation normalization.
Temporal features, including elapsed time and the current temporal target, are excluded from running-statistic normalization and are instead scaled by $T^{\max}$.
Rewards are rescaled using per-task statistics of a backward-discounted reward trace, $z_t=\gamma z_{t-1}+r_t$, maintained separately for each environment and reset after each episode. Each reward is divided by the trace\textquotesingle s running standard deviation. This rescales individual rewards without subtracting a mean; it does not directly normalize forward returns. Relabeled replay rewards use the current on-policy scale without updating its statistics.
When value normalization is enabled, PPO returns, critic predictions, and relabeled replay-buffer returns are mapped into the same normalized value space before value losses or the positive return-value gap are computed.
PPO advantages are normalized per task before the policy gradient update.
For replay actor updates, the positive weights $w_i(\tau)A_t^+(\tau)$ are divided by the standard deviation of their positive entries (without mean subtraction) and capped at $1$; this normalization does not rescale the replay critic loss.

\textbf{Optimization hyperparameters.}
The PPO update, rollout length, minibatch size, discounting, GAE parameter, value coefficient, entropy coefficient, and learning rate schedule are shared with all baselines.
The method-specific settings are the replay buffer size, replay sampling budget, and auxiliary loss coefficients $\lambda_{\mathrm{SI}}$ and $\lambda_{\mathrm{V,SI}}$.
The complete shared PPO settings and replay-specific settings are listed in Tables~\ref{tab:appendix_training_config} and~\ref{tab:appendix_sil_config}.

\section{Training and Evaluation Setup}
\label{app:implementation}

\subsection{Benchmark and agent setup}
\label{app:benchmark}

\begin{figure}[p]
    \centering
    \includegraphics[width=\linewidth]{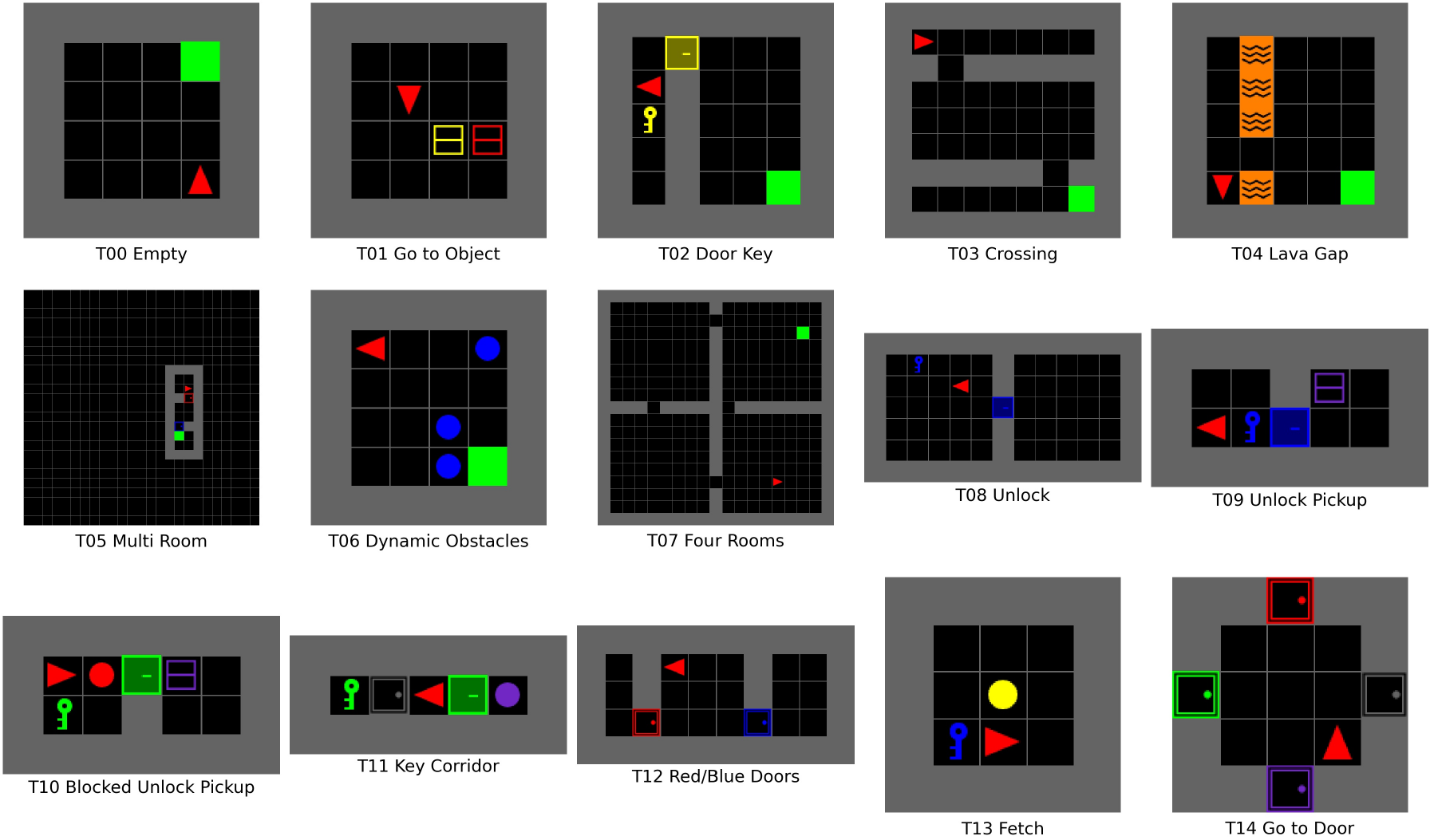}
    \caption{\textbf{MiniGrid task suite.} Representative layouts for the 15 MiniGrid tasks used in our evaluation, covering navigation, obstacle avoidance, door-and-key interaction, object pickup, and goal-conditioned object interaction.}
    \phantomsection
    \label{fig:appendix_minigrid_task_contact_sheet}
    \par\medskip
    \includegraphics[width=\linewidth]{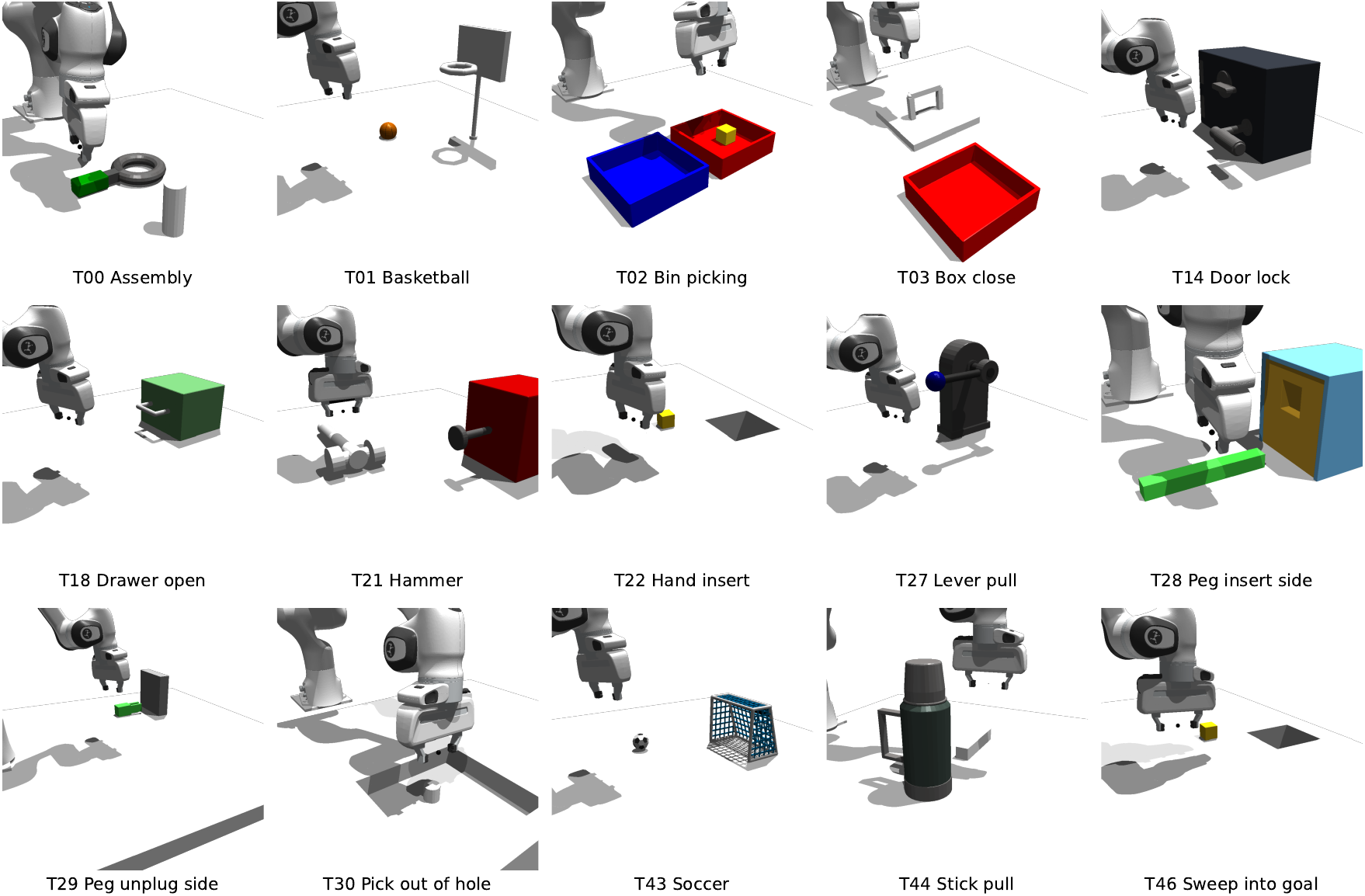}
    \caption{\textbf{Meta-World task suite.} Representative configurations for all 15 Meta-World manipulation tasks.}
    \phantomsection
    \label{fig:appendix_task_contact_sheet}
\end{figure}

\textbf{Tasks.}
We evaluate each method on 15 MiniGrid tasks and 15 Meta-World robot manipulation tasks, with an independently trained policy for each task and three random seeds per task. MiniGrid contains Empty, GoToObject, DoorKey, SimpleCrossing, LavaGap, MultiRoom, DynamicObstacles, FourRooms, Unlock, UnlockPickup, BlockedUnlockPickup, KeyCorridor, RedBlueDoors, Fetch, and GoToDoor. The Meta-World task ids are $\{0,1,2,3,14,18,21,22,27,28,29,30,43,44,46\}$. Figures~\ref{fig:appendix_minigrid_task_contact_sheet} and~\ref{fig:appendix_task_contact_sheet} show the MiniGrid and Meta-World task suites, respectively.

\textbf{MiniGrid agent design.} The MiniGrid policy receives the local $7\times7$ symbolic image and compact mission-target features. Separate actor and critic encoders use three stride-one, unpadded $2\times2$ convolutions with 16, 32, and 64 channels. The first two convolutions use ReLU and the third uses tanh, producing $64\times4\times4=1024$ image features. These features are concatenated with the auxiliary inputs and passed to five-hidden-layer MLPs with widths $[512,256,128,64,32]$, tanh activations, and LayerNorm. The actor outputs categorical action logits and the critic outputs a scalar value. Temporal-target methods additionally receive elapsed steps and the active temporal target, while these inputs are masked for non-temporal baselines.

\textbf{Meta-World agent design.}
The policy observes robot proprioceptive features and task state, including end-effector position, gripper openness, object poses, target position, and a one-step memory of robot and object state.
For temporal-target methods, this observation is augmented with elapsed time and the adaptive temporal target as described in Sec.~\ref{sec:method_temporal_target}.
The action is a four-dimensional continuous command consisting of a Cartesian end-effector position delta and a gripper command, which is executed by the Franka controller in Isaac Gym.
The actor and critic are five-hidden-layer MLPs with hidden sizes $[512,256,128,64,32]$, tanh activations, and LayerNorm.
The actor predicts a mean $\mu_\theta(\tilde{o}_t)$ and uses a state-independent diagonal covariance,
\begin{equation}
    a_t \sim \pi_\theta(\cdot\mid\tilde{o}_t),
    \qquad
    \pi_\theta(\cdot\mid\tilde{o}_t)
    =
    \mathcal{N}\!\left(\mu_\theta(\tilde{o}_t),\operatorname{diag}(\sigma^2)\right),
\end{equation}
where $\sigma\in\mathbb{R}^4$ is the action-standard-deviation vector parameterized by a learned state-independent log-standard-deviation vector.
The operator $\operatorname{diag}(\sigma^2)$ forms a diagonal covariance matrix with entries $\sigma_j^2$ on the diagonal. The policy observation can be obtained on a real setup from robot state estimation and calibrated perception or tracking. Elapsed time and the temporal target are directly available scalar signals.

\phantomsection\textbf{Meta-World controller design.}\label{app:controller}
The high-level policy outputs Cartesian end-effector position deltas and a gripper command at 60~Hz.
The arm command is converted by damped differential inverse kinematics into target joint positions.
These joint targets can be tracked by a low-level joint impedance controller at 250~Hz, which converts joint-position errors into torque commands.
The differential IK module limits arm-joint commands using nominal velocity bounds of $2.175$~rad/s for joints 1--4 and $2.610$~rad/s for joints 5--7, and includes an exponential moving average smoothing option for joint increments, which can be used for future real-robot deployment to attenuate abrupt command changes and reduce oscillatory motion.
This hierarchy keeps the learned interface close to standard real-robot control stacks by delegating robot-specific dynamics to IK and impedance control.

\subsection{Disturbed training settings}
\label{app:disturbance_settings}

Disturbance experiments compare SIL, ATTL, and TSIL while perturbing one training factor and leaving the remaining preprocessing and loss-scaling settings unchanged. For MiniGrid, we evaluate policy-gradient noise and increased PPO clip ratios. For Meta-World, we additionally evaluate dense-reward dropout. Table~\ref{tab:appendix_disturbance_settings} summarizes the applicable disturbance classes.

\textbf{Policy gradient noise.} We add Gaussian noise to the PPO actor gradient and restore its original global norm before combining it with the other gradients:
\begin{equation}
    \begin{aligned}
        \epsilon_{\alpha,b} &= \alpha_{\mathrm{noise}}\operatorname{RMS}(g_{\mathrm{PPO},b})z_b,
        \qquad z_b\sim\mathcal{N}(0,I), \\
        g_{\mathrm{actor}} &=
        \frac{\lVert g_{\mathrm{PPO}}\rVert_2}{\lVert g_{\mathrm{PPO}}+\epsilon_\alpha\rVert_2}
        (g_{\mathrm{PPO}}+\epsilon_\alpha)+g_{\mathrm{replay}}+g_{\mathrm{other}}.
    \end{aligned}
    \label{eq:appendix_gradient_noise}
\end{equation}
Here $b$ indexes a parameter tensor, $\alpha_{\mathrm{noise}}\in\{5,10,20\}$ is the dimensionless noise scale, and $\operatorname{RMS}(g)=\sqrt{\sum_{j=1}^{d}g_j^2/d}$ for a tensor with $d$ entries. Noise entries are sampled independently; the norms span all PPO actor parameters. The PPO gradient includes entropy regularization, $g_{\mathrm{replay}}$ includes the replay coefficient, and $g_{\mathrm{other}}$ contains other actor regularization gradients. Replay and other gradients receive no injected noise. If norm restoration is undefined, we retain the clean PPO gradient. The combined gradient undergoes the usual clipping and optimizer update.

\textbf{Dense reward dropout.}
We dropped dense shaping rewards episode-wise with dropout probability $p$, while keeping the target-conditioned success reward.
The dropout mask was resampled when an environment finished an episode.
This tested whether replay could help policy training when shaped guidance became intermittent or unreliable.

\textbf{Large PPO clip ratio.}
We increased the PPO clip ratio value beyond the default setting, allowing larger policy updates per minibatch and weakening PPO's trust-region-style constraint.
This sweep tested whether self-imitation could stabilize these more aggressive updates while preserving training efficiency.

\begin{table}[h]
    \centering
    \caption{\textbf{Disturbed training settings.} Each experiment varies one disturbance type while keeping the remaining training configuration fixed.}
    \label{tab:appendix_disturbance_settings}
    \footnotesize
    \setlength{\tabcolsep}{4pt}
    \begin{tabular}{lll}
        \toprule
        Disturbance type & MiniGrid values & Meta-World values \\
        \midrule
        Policy gradient noise scaling & $5,10,20$ & $5,10,20$ \\
        Dense reward dropout probability & --- & $0.4,0.6,0.8$ \\
        PPO clip ratio & $0.3,0.5,0.7$ & $0.3,0.5,0.7$ \\
        \bottomrule
    \end{tabular}
\end{table}

\subsection{Baselines and hyperparameters}
\label{app:baselines}

Within each benchmark, all methods use PPO as the base optimizer and share the same agent architecture, rollout collection, preprocessing, and other training components.
We used the baseline set described in Sec.~\ref{sec:experiments}.
All methods use the same base success-reward coefficient within each benchmark: $r_{\mathrm{succ}}=100$.
Tables~\ref{tab:appendix_minigrid_training_config} and~\ref{tab:appendix_training_config} list the benchmark-specific training configurations.

Self-imitation methods use the same auxiliary coefficients and perform joint updates every PPO minibatch. Table~\ref{tab:appendix_sil_config} lists these replay-specific settings.

\begin{table}[h]
    \centering
    \caption{\textbf{MiniGrid training configuration.} Episode horizons are fixed separately for each task.}
    \label{tab:appendix_minigrid_training_config}
    \footnotesize
    \setlength{\tabcolsep}{4pt}
    \begin{tabular}{ll}
        \toprule
        Item & Value \\
        \midrule
        Parallel environments & $32$ \\
        Episode horizon & $100$--$720$ steps, task dependent \\
        Total training budget & $640{,}000$ steps \\
        PPO rollout length & $32$ steps \\
        Discount / GAE parameter & $\gamma=0.995$ / $\lambda=0.95$ \\
        Learning rate & $5\times10^{-4}$ \\
        Success reward scale & $r_{\mathrm{succ}}=100$ \\
        PPO clip ratio & $0.2$ \\
        PPO epochs / minibatch size & $5$ / $128$ \\
        Value coefficient & $4$ \\
        Entropy coefficient & $0.005$ \\
        Evaluation trials per task per seed & $2000$ \\
        Evaluation configurations per task per seed & $32$ \\
        Evaluation checkpoint & best final-$10\%$ training success rate \\
        \bottomrule
    \end{tabular}
\end{table}

\begin{table}[h]
    \centering
    \caption{\textbf{Meta-World training configuration.} Values apply unless explicitly swept in the disturbance experiments.}
    \label{tab:appendix_training_config}
    \footnotesize
    \setlength{\tabcolsep}{4pt}
    \begin{tabular}{ll}
        \toprule
        Item & Value \\
        \midrule
        Parallel environments & $6400$ \\
        Episode horizon & $150$ steps \\
        Total training budget & $50\times10^6$ steps \\
        PPO rollout length & $32$ steps \\
        Discount / GAE parameter & $\gamma=0.995$ / $\lambda=0.95$ \\
        Learning rate & $5\times10^{-4}$ \\
        Success reward scale & $r_{\mathrm{succ}}=100$ \\
        PPO clip ratio & $0.2$ \\
        PPO epochs / minibatch size & $5$ / $20480$ \\
        Value coefficient & $4$ \\
        Entropy coefficient & $0.005$ \\
        Evaluation trials per task per seed & $2000$ \\
        Evaluation configurations per task per seed & $2000$ \\
        Evaluation checkpoint & best final-$10\%$ training success rate \\
        \bottomrule
    \end{tabular}
\end{table}

\begin{table}[b]
    \centering
    \caption{\textbf{SIL-specific replay configuration.} SIL stores high-return trajectories, while TSIL prioritizes fast successful trajectories and fills unused replay slots with high-return trajectories. Both methods use the same gap-proportional sampler.}
    \label{tab:appendix_sil_config}
    \footnotesize
    \setlength{\tabcolsep}{3pt}
    \begin{adjustbox}{max width=\linewidth}
    \begin{tabular}{lll}
        \toprule
        Item & SIL & TSIL \\
        \midrule
        Replay source & high-return & fast-success first with high-return as fallback \\
        Sample selection & positive-gap proportional, without replacement & positive-gap proportional, without replacement \\
        Trajectory weighting & none & efficiency weighting \\
        Per-env. replay buffer top-$k$ & $5$ & $5$ \\
        Maximum sampled transitions per selected trajectory, $n$ & $32$ & $32$ \\
        Replay samples per PPO update & $\leq k n N_{\mathrm{env}}$ & $\leq k n N_{\mathrm{env}}$ \\
        Policy replay coefficient & $\lambda_{\mathrm{SI}}=0.05$ & $\lambda_{\mathrm{SI}}=0.05$ \\
        Value replay coefficient & $\lambda_{\mathrm{V,SI}}=0.05$ & $\lambda_{\mathrm{V,SI}}=0.05$ \\
        \bottomrule
    \end{tabular}
    \end{adjustbox}
\end{table}

\subsection{Primary metric computation}
\label{app:metrics}

Training success rate is the maximum rolling success rate recorded during the final 10\% of the training budget. Training AUC is the trapezoidal integral of the rolling success curve over the full interaction budget, divided by that budget. Successful episode count is the total number of successful episodes collected during training.

For an episode $e$ with duration $T_e$, success indicator $s_e$, and fixed task horizon $T_e^{\max}$, normalized completion time is $s_e T_e/T_e^{\max}+(1-s_e)$. During training, we average this quantity over the same rolling window used for success rate and report its value at the checkpoint with the maximum rolling success rate within the final 10\% of training. Evaluation success rate is computed from the reported policy using the benchmark evaluation protocol and the checkpoint with the maximum final-10\% training success rate.

\textbf{Rolling statistics and evaluation actions.} The rolling success rate averages the most recent completed episodes of each task, using a buffer of $\max(5N_{\mathrm{env}},50)$ episodes. With independently trained policies, this gives 160 episodes for MiniGrid and 32,000 for Meta-World. Evaluation samples actions from the learned categorical policy in MiniGrid and Gaussian policy in Meta-World.

\raggedbottom
\section{Additional Training Results}
\label{app:additional_results}
\begin{figure}[H]
    \centering
    \includegraphics[width=\linewidth]{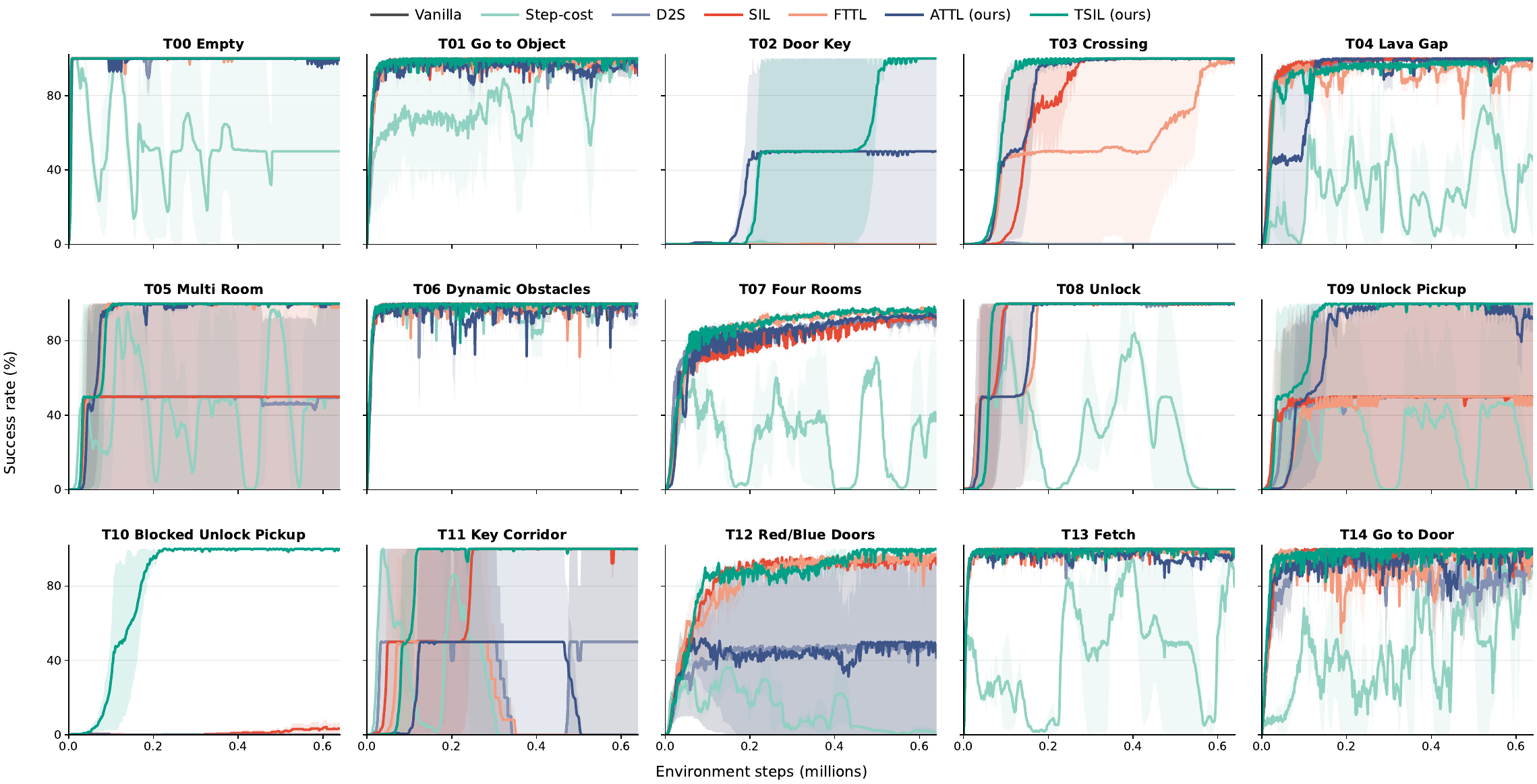}
    \caption{\textbf{Per-task MiniGrid success rate.} Success rate over training for all 15 MiniGrid tasks. Higher values indicate more reliable task completion.}
    \label{fig:appendix_minigrid_success_curves}
\end{figure}

\begin{figure}[H]
    \centering
    \includegraphics[width=\linewidth]{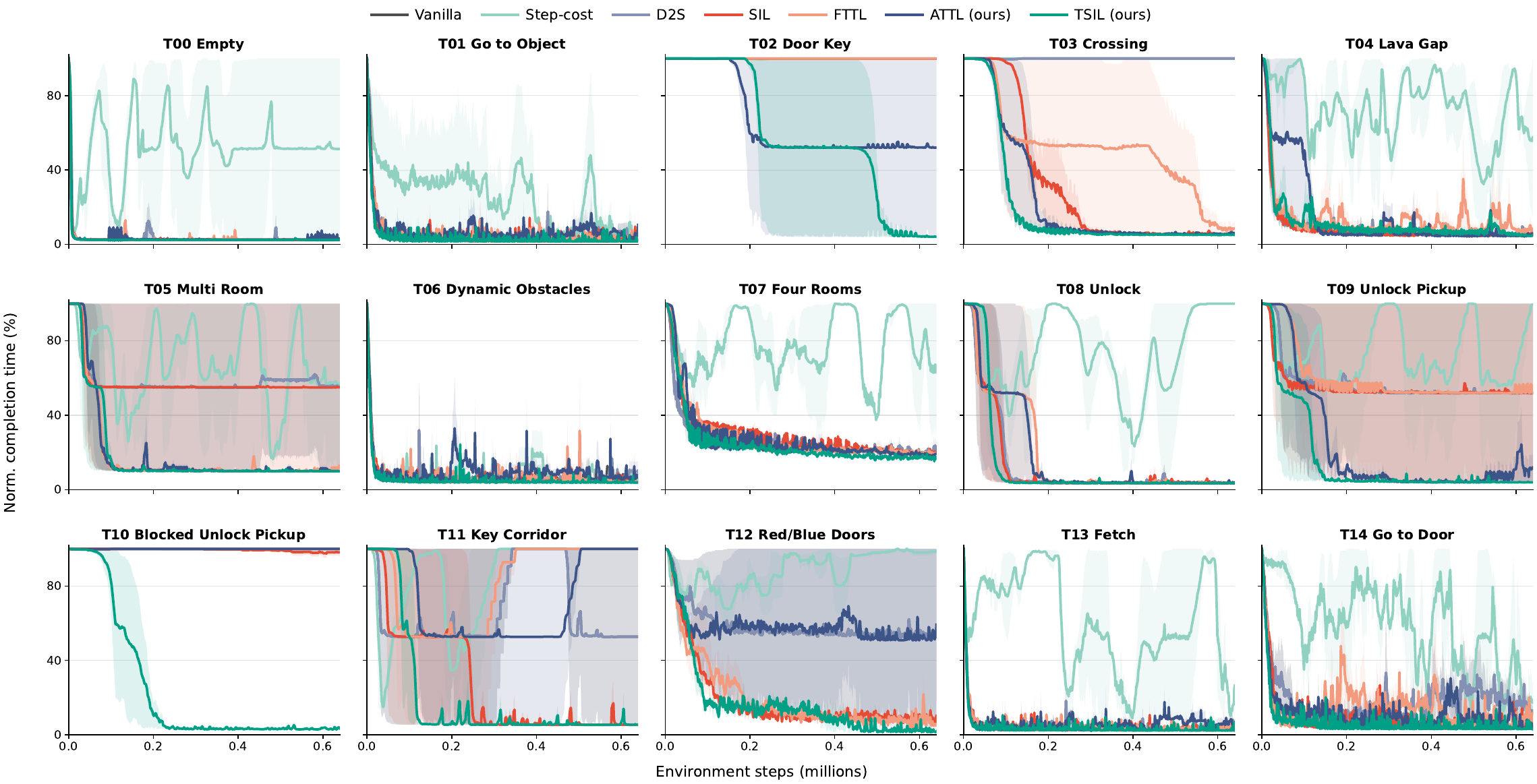}
    \caption{\textbf{Per-task MiniGrid normalized completion time.} Normalized completion time over training for all 15 MiniGrid tasks. Failures receive the full task horizon, and lower values indicate more reliable and temporally efficient completion.}
    \label{fig:appendix_minigrid_time_curves}
\end{figure}

\begin{figure}[H]
    \centering
    \includegraphics[width=\linewidth]{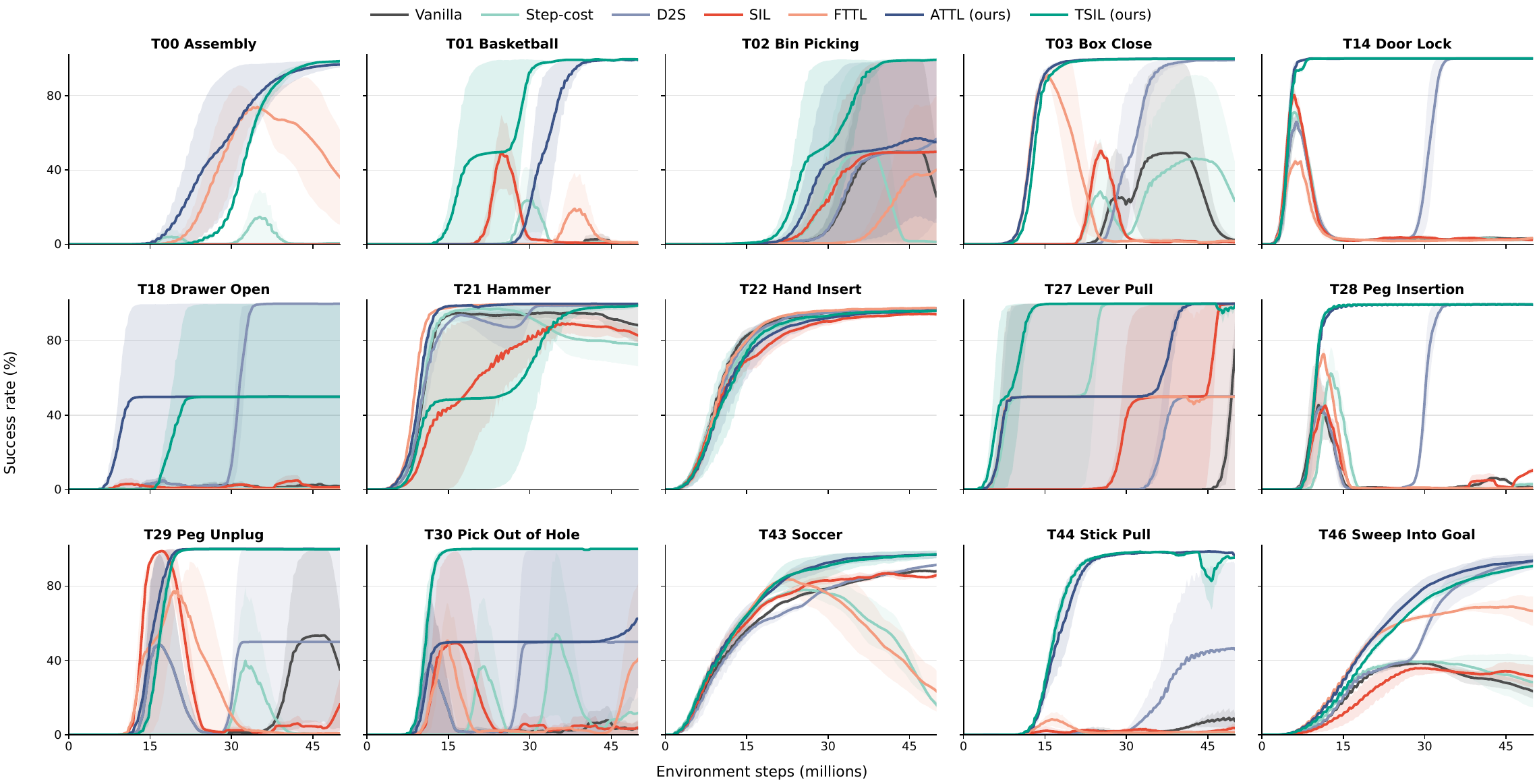}
    \caption{\textbf{Per-task Meta-World success rate.} Success rate over training for all 15 Meta-World tasks. Higher values indicate that the policy completes the task more reliably.}
    \label{fig:appendix_metaworld_success_curves}
\end{figure}

\begin{figure}[H]
    \centering
    \includegraphics[width=\linewidth]{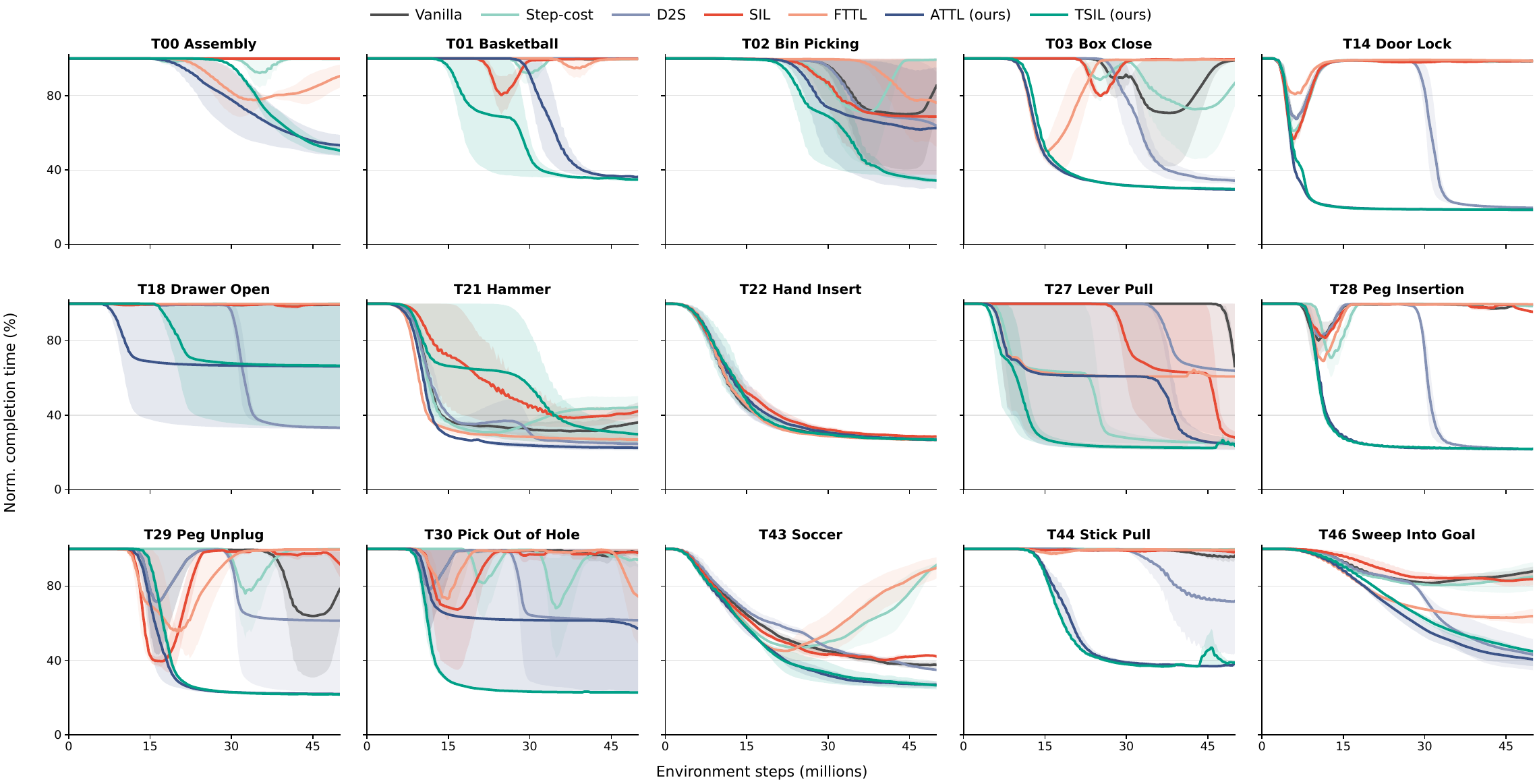}
    \caption{\textbf{Per-task Meta-World normalized completion time.} Normalized completion time over training for all 15 Meta-World tasks. Failures receive the full task horizon, and lower values indicate more reliable and temporally efficient completion.}
    \label{fig:appendix_metaworld_time_curves}
\end{figure}

Figures~\ref{fig:appendix_minigrid_success_curves}--\ref{fig:appendix_metaworld_time_curves} expand main-text Fig.~\ref*{fig:main_results} with standard-training learning curves for every task. Lines show means across seeds, and shaded regions show $\pm$ one standard deviation (SD) across seeds.
\subsection{MiniGrid}

\clearpage
\subsection{Meta-World}

\clearpage
\section{Additional Analysis}
\label{app:analysis}
\subsection{Learning-signal metric computation}
\label{app:diagnostic_metrics}
We provide detailed definitions for the diagnostic metrics used in Sec.~\ref{sec:exp_analysis}.
These metrics quantify where positive-advantage mass concentrates and how self-imitation affects revisitation of stored behavior.

\textbf{Learning-signal diagnostics.}
For the learning-signal analysis, completed episodes were grouped into bins by normalized completion time and discounted task-reward return. As shown in Fig.~\ref{fig:addl_signal}, cell color shows empirical success rate in each bin, while the red overlay gradient shows the fraction of total positive-advantage mass assigned to that bin.
Let $\mathcal{E}$ be the set of completed training episodes, and let $b\subset\mathcal{E}$ denote the subset of episodes assigned to a given visualization bin.
We define
\begin{equation}
    \mathrm{Mass}(b)
    =
    \frac{
        \sum_{\tau\in b}\sum_{t\in\tau} [\hat{A}^{\mathrm{PPO}}_t(\tau)]_+
    }{
        \sum_{\tau\in\mathcal{E}}\sum_{t\in\tau} [\hat{A}^{\mathrm{PPO}}_t(\tau)]_+
    },
    \label{eq:positive_advantage_mass}
\end{equation}
where $[x]_+=\max(x,0)$ and $\hat{A}^{\mathrm{PPO}}_t(\tau)$ is the advantage after the same normalization used for the PPO policy update.

For the summary bars on the right, let $m_\tau=\sum_{t\in\tau}[\hat{A}^{\mathrm{PPO}}_t(\tau)]_+$ be the positive-advantage mass of episode $\tau$, and let $M=\sum_{\tau\in\mathcal{E}}m_\tau$ be the total positive-advantage mass over the completed episodes.
We also let $s_\tau\in\{0,1\}$ be the success indicator, $\bar{T}_\tau=T_\tau/T^{\max}$ be normalized completion time, and $G_\tau^{\mathrm{task}}=\sum_{t\in\tau}\gamma^t r_t^{\mathrm{task}}$ be the discounted return from the task-reward component.
Let $G_{\min}^{\mathrm{task}}=\min_{\tau\in\mathcal{E}}G_\tau^{\mathrm{task}}$, $G_{\max}^{\mathrm{task}}=\max_{\tau\in\mathcal{E}}G_\tau^{\mathrm{task}}$, and define the normalized task return as
\begin{equation}
    \rho_\tau
    =
    \frac{G_\tau^{\mathrm{task}}-G_{\min}^{\mathrm{task}}}{\max(G_{\max}^{\mathrm{task}}-G_{\min}^{\mathrm{task}},\epsilon)},
    \label{eq:dense_return_rank}
\end{equation}
where $\epsilon$ is a small numerical constant.
The four aggregate mass fractions are defined as:
\begin{equation}
    F_{\mathrm{success}}
    =
    \frac{1}{M}
    \sum_{\tau\in\mathcal{E}}
    m_\tau s_\tau ,
    \label{eq:success_mass_fraction}
\end{equation}
\begin{equation}
    F_{\mathrm{fastsuccess}}
    =
    \frac{1}{M}
    \sum_{\tau\in\mathcal{E}}
    m_\tau s_\tau (1-\bar{T}_\tau),
    \label{eq:fast_success_mass_fraction}
\end{equation}
\begin{equation}
    F_{\mathrm{slowfailure}}
    =
    \frac{1}{M}
    \sum_{\tau\in\mathcal{E}}
    m_\tau (1-s_\tau)\bar{T}_\tau,
    \label{eq:slow_failure_mass_fraction}
\end{equation}
\begin{equation}
    F_{\mathrm{rewarddistracted}}
    =
    \frac{1}{M}
    \sum_{\tau\in\mathcal{E}}
    m_\tau \rho_\tau \bar{T}_\tau,
    \label{eq:reward_distracted_mass_fraction}
\end{equation}
If $M=0$, all fractions are set to zero.
$F_{\mathrm{success}}$ measures how much positive-advantage mass comes from successful episodes, indicating whether the update reinforces task completion.
$F_{\mathrm{fastsuccess}}$ further weights successful episodes by temporal efficiency, indicating whether the update concentrates on earlier successes.
$F_{\mathrm{slowfailure}}$ measures positive-advantage mass assigned to late unsuccessful trajectories, which indicates whether PPO still reinforces horizon-consuming failures.
$F_{\mathrm{rewarddistracted}}$ measures positive-advantage mass assigned to late trajectories with high dense task return, indicating whether shaped rewards attract updates toward slow reward-seeking behavior rather than compact completion.

\clearpage
\subsection{Additional learning-signal diagnostics}
\label{app:learning_signals}
\begin{figure}[H]
    \centering
    \includegraphics[width=\linewidth]{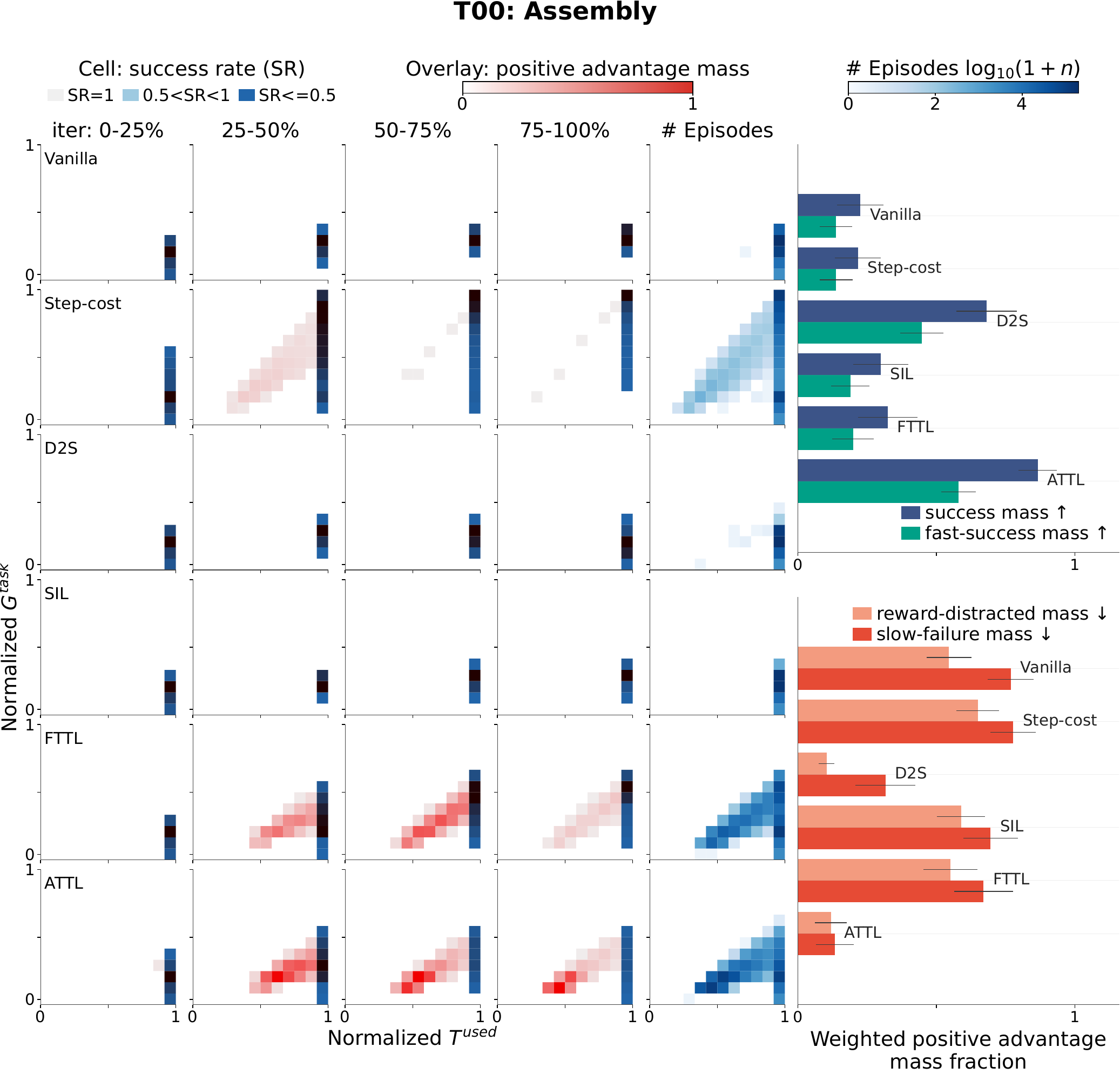}
    \caption{\textbf{Learning-signal analysis for T00: Assembly.} The panels use the same construction and metrics as Fig.~\ref{fig:addl_signal}.}
    \label{fig:appendix_training_signal_t00}
\end{figure}

\begin{figure}[H]
    \centering
    \includegraphics[width=\linewidth]{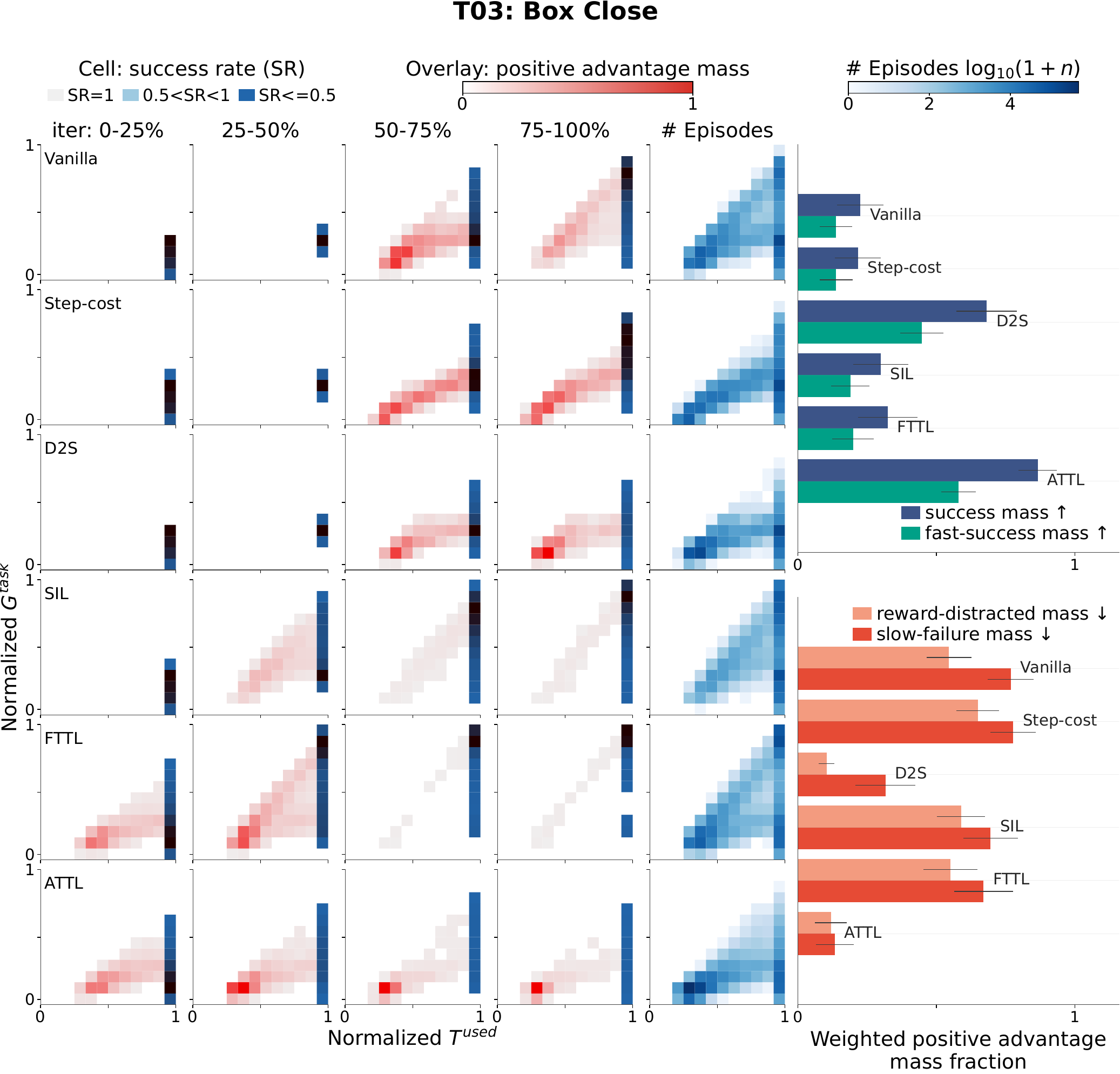}
    \caption{\textbf{Learning-signal analysis for T03: Box Close.} The panels use the same construction and metrics as Fig.~\ref{fig:appendix_training_signal_t00}.}
    \label{fig:appendix_training_signal_t03}
\end{figure}

\begin{figure}[H]
    \centering
    \includegraphics[width=\linewidth]{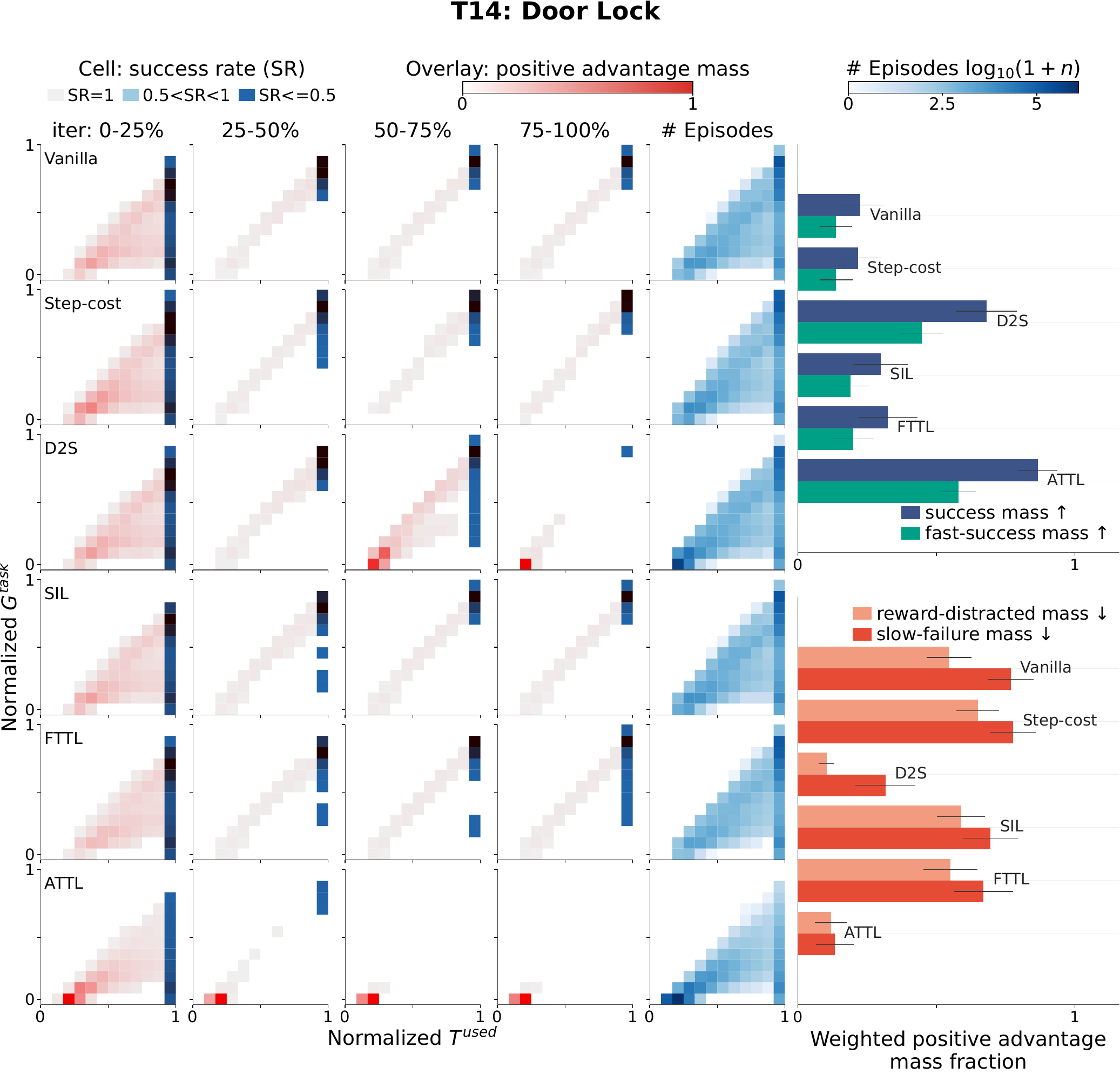}
    \caption{\textbf{Learning-signal analysis for T14: Door Lock.} The panels use the same construction and metrics as Fig.~\ref{fig:appendix_training_signal_t00}.}
    \label{fig:appendix_training_signal_t14}
\end{figure}

\begin{figure}[H]
    \centering
    \includegraphics[width=\linewidth]{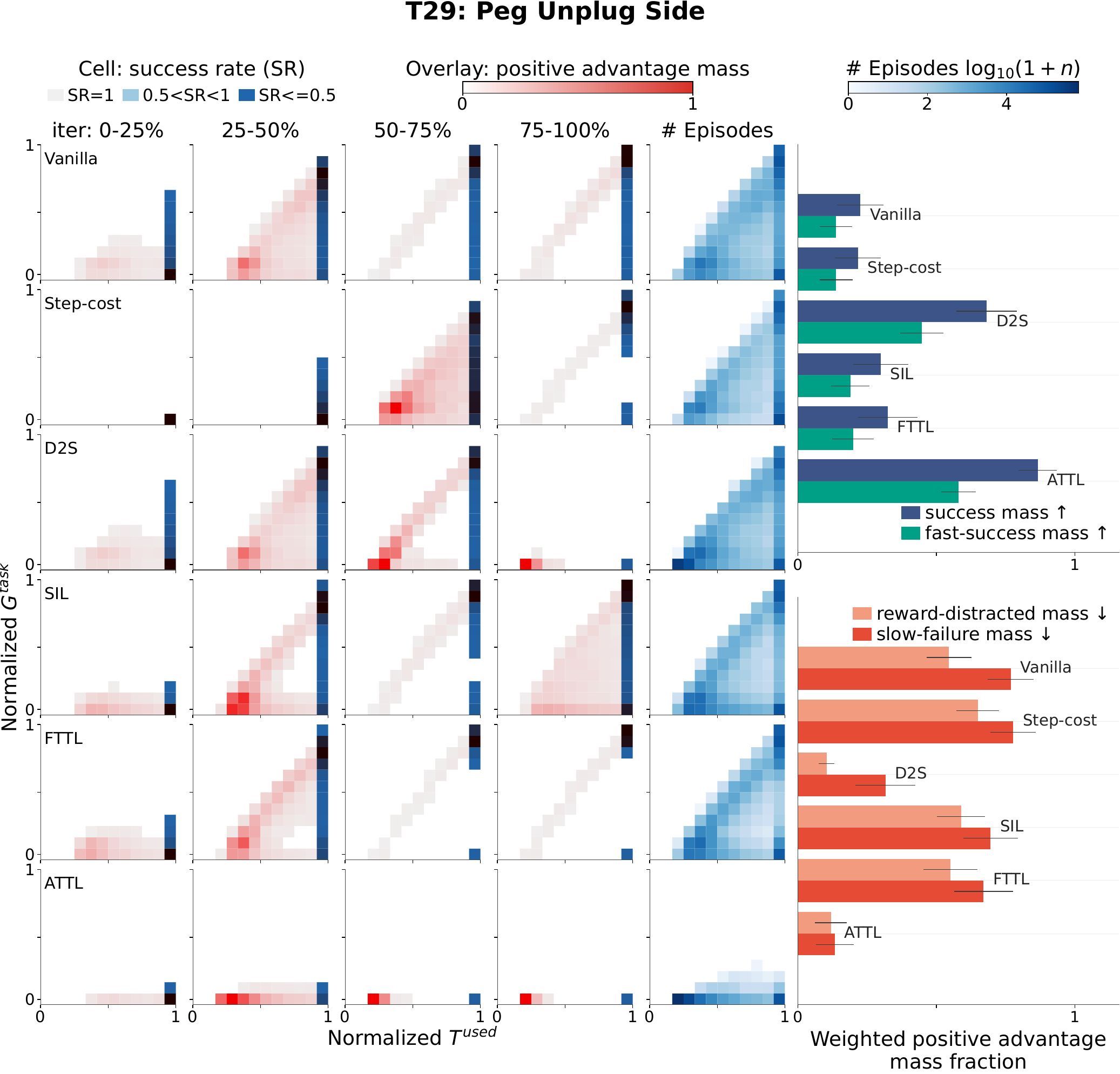}
    \caption{\textbf{Learning-signal analysis for T29: Peg Unplug Side.} The panels use the same construction and metrics as Fig.~\ref{fig:appendix_training_signal_t00}.}
    \label{fig:appendix_training_signal_t29}
\end{figure}

\begin{figure}[H]
    \centering
    \includegraphics[width=\linewidth]{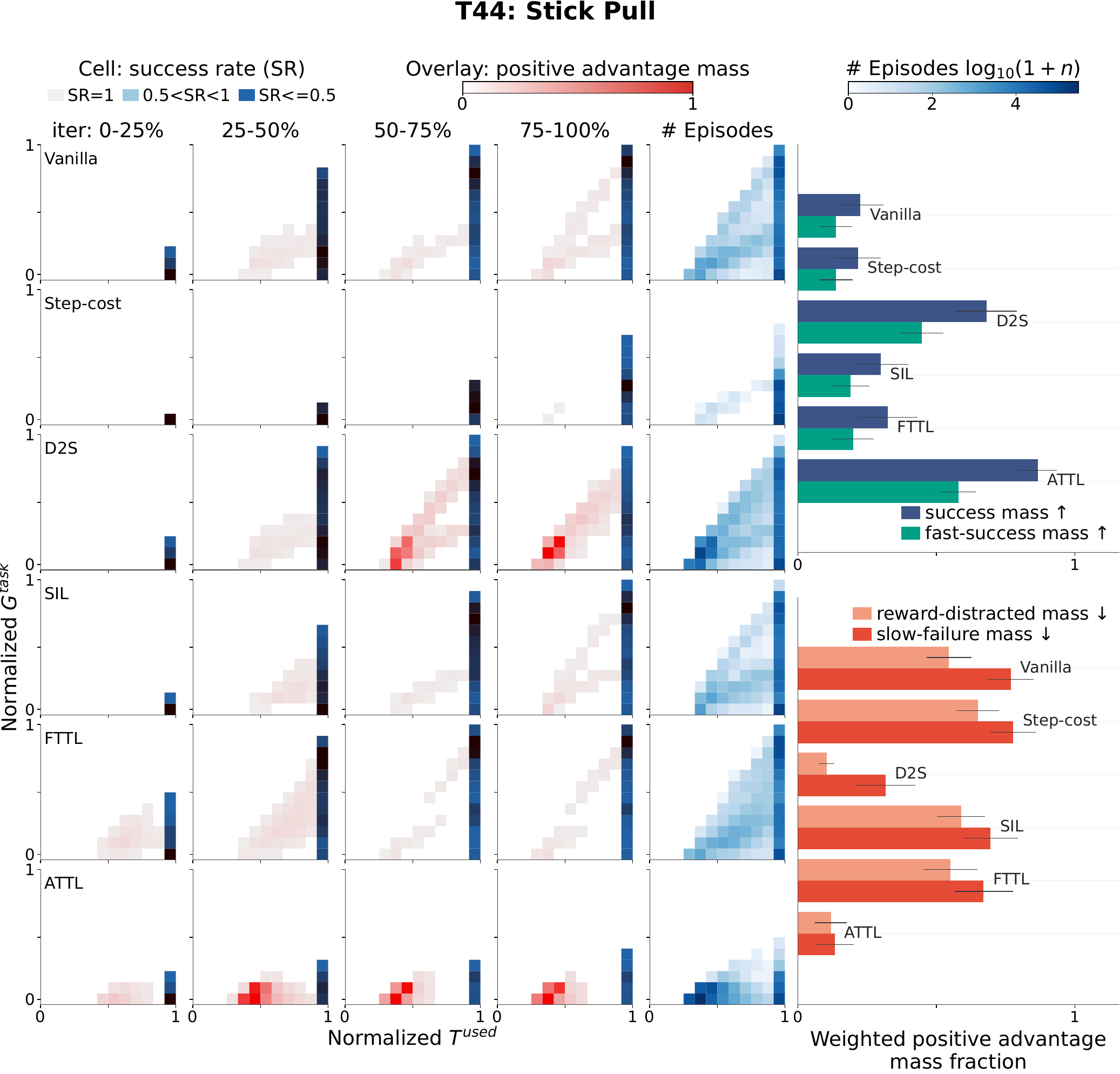}
    \caption{\textbf{Learning-signal analysis for T44: Stick Pull.} The panels use the same construction and metrics as Fig.~\ref{fig:appendix_training_signal_t00}.}
    \label{fig:appendix_training_signal_t44}
\end{figure}

Figures~\ref{fig:appendix_training_signal_t00}--\ref{fig:appendix_training_signal_t44} extend the analysis in main-text Fig.~\ref*{fig:addl_signal} to five additional Meta-World tasks. Each row is a method; the first four columns divide training into four equal interaction intervals. Cell color shows empirical success rate over normalized completion time and normalized task return, while the red overlay shows the fraction of positive-advantage mass. The episode-count map and summary bars show how each method distributes its update signal across successful, fast-successful, reward-distracted, and slow-failure behavior; metric definitions are in Sec.~\ref{app:diagnostic_metrics}.
\clearpage
\textbf{Learning-signal diagnostics (continued).}
\clearpage
\textbf{Learning-signal diagnostics (continued).}
\clearpage
\textbf{Learning-signal diagnostics (continued).}
\clearpage
\textbf{Learning-signal diagnostics (continued).}

\clearpage
\subsection{Policy-update landscape construction}
\label{app:landscape_method}
Main-text Fig.~\ref*{fig:replay_landscape} plots mean action log probability on fixed fast-success memory, averaged first over sampled transitions per trajectory, then over trajectories. The horizontal axis follows the negative PPO policy-loss gradient; the vertical axis follows the orthogonal component of the memory log-probability gradient, both normalized. SIL records this memory only for diagnostics, not replay. Surfaces evaluate parameter perturbations in this plane; normalized PPO and replay-combined directions are projected as arrows indicating direction, not step size. Each surface is independently rescaled to $[0,1]$, so colors indicate within-panel relative likelihood. Black points mark current policies; percentages report rollout success. Panels show 25\%, 50\%, 75\%, and 100\% of training.

\subsection{Additional policy-update landscapes}
\label{app:policy_landscapes}
These plots extend main-text Fig.~\ref*{fig:replay_landscape}. Each task occupies one row: the first four panels show PPO and TSIL updates, and the last four show SIL updates. These landscapes suggest that TSIL better preserves fast-success behavior. This improves learning efficiency by making better use of scarce early fast-successes and stabilizes training by reducing the forgetting of efficient successful behaviors.
\begin{figure}[H]
    \centering
    \includegraphics[width=\linewidth]{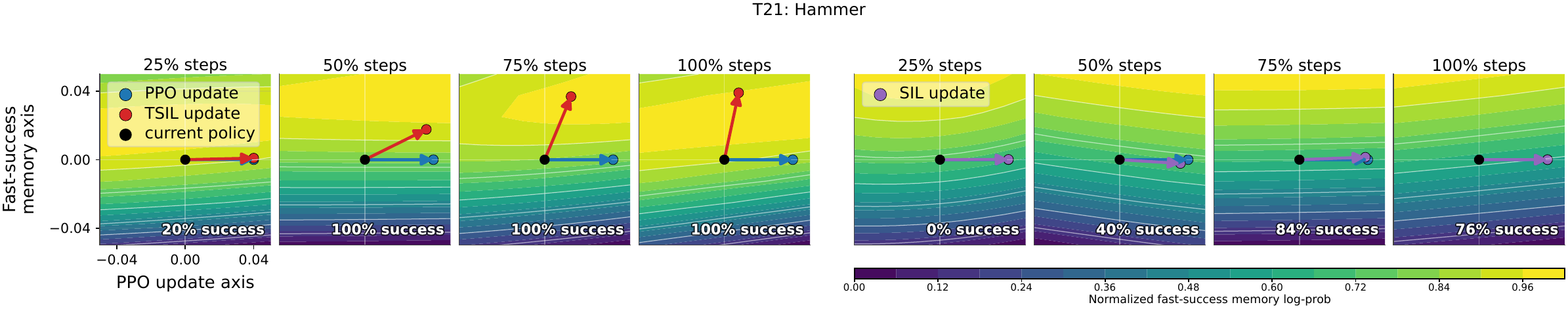}\par\smallskip
    \includegraphics[width=\linewidth]{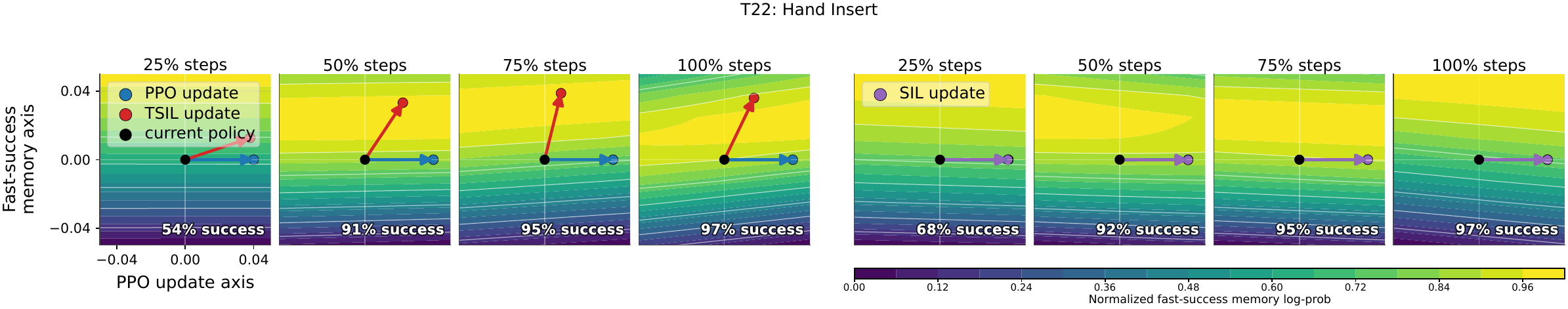}\par\smallskip
    \includegraphics[width=\linewidth]{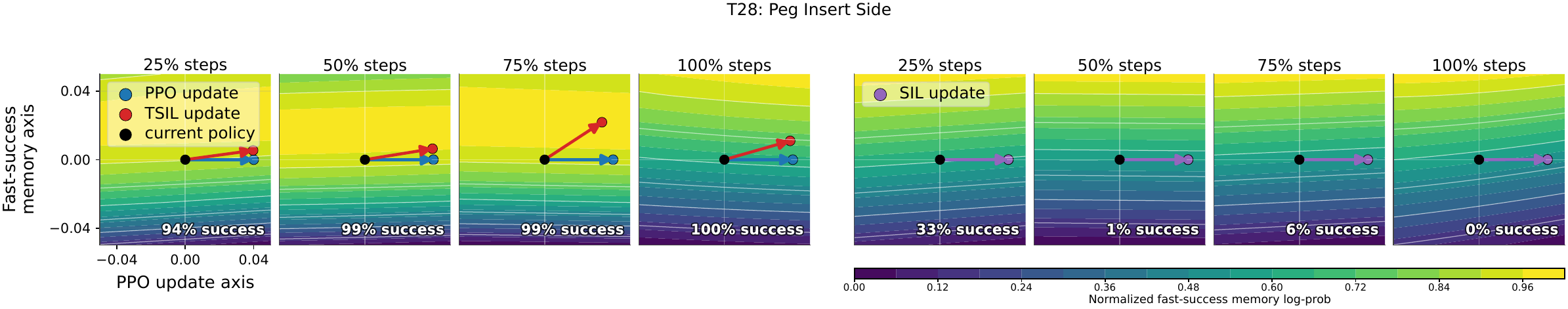}\par\smallskip
    \includegraphics[width=\linewidth]{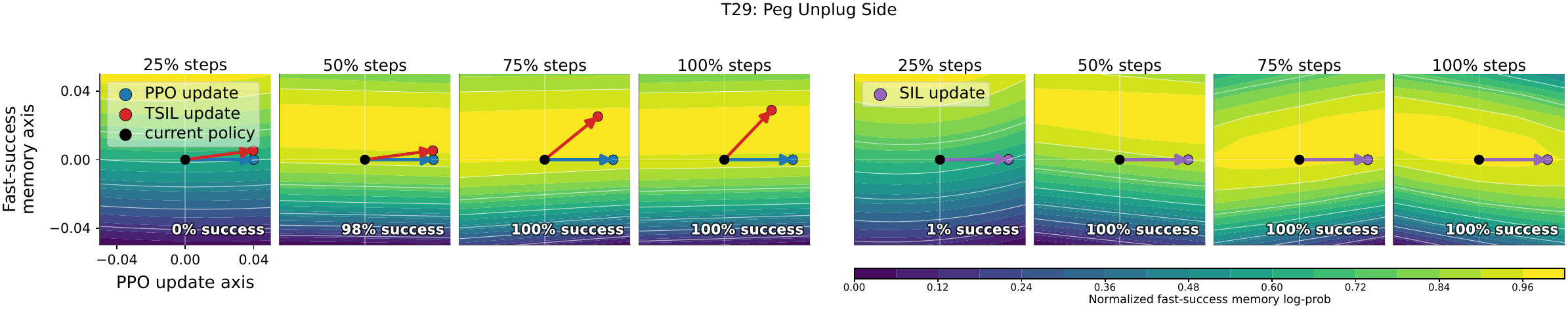}\par\smallskip
    \includegraphics[width=\linewidth]{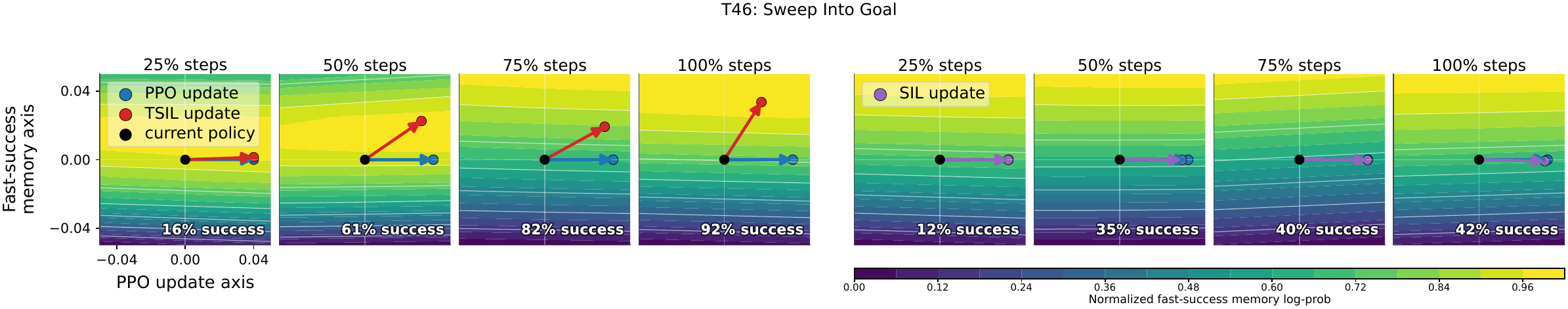}
    \caption{\textbf{Additional policy-update landscapes.} Each group of four panels shows 25\%, 50\%, 75\%, and 100\% of training. The landscapes suggest stronger preservation of fast-success behavior with TSIL than with SIL.}
    \label{fig:appendix_policy_landscapes}
\end{figure}
\clearpage

\end{document}